\pdfoutput=1

\documentclass[11pt]{article}

\usepackage[final]{acl}

\usepackage{times}
\usepackage{latexsym}

\usepackage[T1]{fontenc}

\usepackage[utf8]{inputenc}

\usepackage{microtype}

\usepackage{inconsolata}

\usepackage{graphicx}

\usepackage{kotex}
\usepackage{amssymb}
\usepackage{amsmath}
\usepackage{booktabs}
\usepackage{geometry}
\usepackage{tcolorbox}
\usepackage{lipsum} 

\usepackage{colortbl}
\definecolor{headercolor}{HTML}{E0EBF6}
\definecolor{conceptcolor}{HTML}{E5F0DB} 
\definecolor{answerbackcolor}{HTML}{F7E6D8} 

\title{Towards Scalable Lifelong Knowledge Editing with Selective Knowledge Suppression}

\author{
  Dahyun Jung \quad Jaewook Lee \quad Heuiseok Lim\thanks{Corresponding author.} \\
  Department of Computer Science and Engineering, Korea University \\
  \texttt{\{dhaabb55,jaewook133,limhseok\}@korea.ac.kr}
}

\begin{document}
\maketitle
\begin{abstract}
Large language models (LLMs) require frequent knowledge updates to reflect changing facts and mitigate hallucinations. To meet this demand, lifelong knowledge editing has emerged as a continual approach to modify specific pieces of knowledge without retraining the entire model. Existing parameter-editing methods struggle with stability during sequential edits due to catastrophic forgetting. While retrieval-based approaches are proposed to alleviate this issue, their applicability remains limited across various datasets because of high training costs. To address these limitations and enhance scalability in lifelong settings, we propose LightEdit. Our framework first selects relevant knowledge from retrieved information to modify the query effectively. It then incorporates a decoding strategy to suppress the model’s original knowledge probabilities, thereby enabling efficient edits based on the selected information. Extensive experiments on ZSRE, Counterfact, and RIPE benchmarks demonstrate that LightEdit outperforms existing lifelong knowledge editing methods. Furthermore, by minimizing training costs, LightEdit achieves cost-effective scalability, enabling easy adaptation to various datasets.\footnote{Our code is available at \url{https://github.com/ekgus9/LightEdit}.}
\end{abstract}

\section{Introduction}
Large language models~(LLMs) have demonstrated impressive performance across a wide range of natural language processing~(NLP) tasks, primarily attributed to their pretraining on large-scale corpora~\cite{alabdulmohsin2022revisiting,jiang2023mistral,haviv2023understanding,openai2023gpt4,zhao2025surveylargelanguagemodels}. However, LLMs continue to exhibit limitations (e.g., hallucination and bias). They struggle to incorporate continually evolving knowledge due to their reliance on static pre-training data~\cite{pagnoni2021understanding, Ji_2023, shi2024continuallearninglargelanguage}. While retraining offers a potential solution, it remains prohibitively expensive in terms of computational resources~\cite{lazaridou2021mind}.

Knowledge editing is proposed to modify specific knowledge of LLMs without costly retraining~\cite{mitchell2022memorybased,yao2023editinglargelanguagemodels,pinter2023emptyingoceanspoonedit,wang2023knowledge,zhang2024comprehensive}. Existing knowledge editing studies mainly focus on scenarios involving editing a single or multiple knowledge entries simultaneously~\cite{meng2023massediting,tan2024massiveeditinglargelanguage,sharma2024locatingeditingfactualassociations,xie2024memlaenhancingmultilingualknowledge,jiang2025anyediteditknowledgeencoded}. However, real-world applications require continual updates to knowledge, necessitating consideration of lifelong knowledge editing scenarios~\cite{huang2023transformerpatchermistakeworthneuron,hartvigsen2023aginggracelifelongmodel,hu2024wilkewiselayerknowledgeeditor}.

\begin{figure*}[hbt!]
\centering 
\includegraphics[width=1\linewidth]{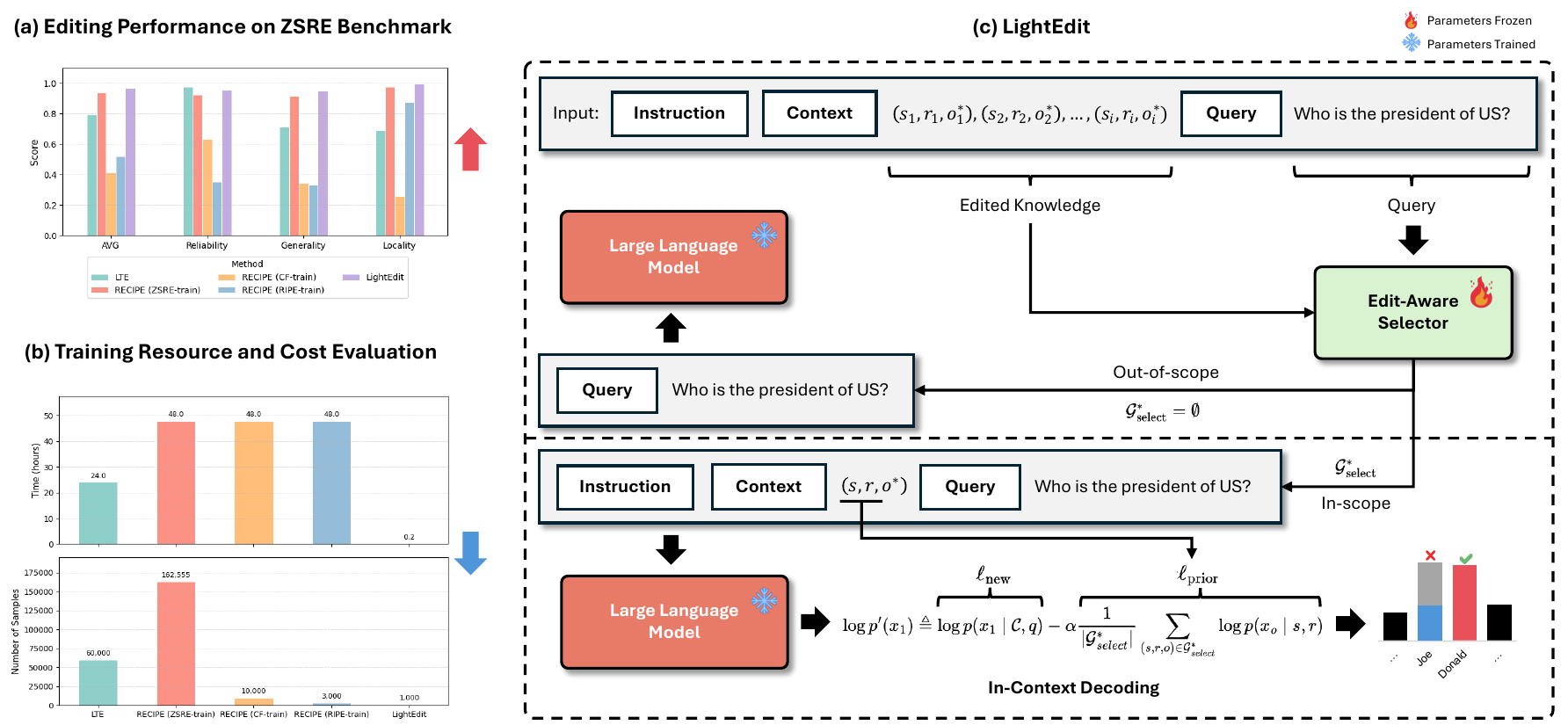}
\caption{Experimental comparison with previous lifelong knowledge editing methods and the overall framework of LightEdit.
(a) demonstrates that LightEdit significantly improves editing performance, while RECIPE suffers from substantial performance degradation when evaluated on datasets different from those used during training. (b) shows that LightEdit is also more efficient in terms of training time and data volume than existing methods. These advantages arise from the architecture illustrated in (c). We employ an edit-aware selector to retrieve the knowledge necessary for query editing. When modified knowledge is selected, we mitigate potential knowledge conflicts during inference by applying in-context decoding, which downweights the generation probability of outdated knowledge.
}
\label{fig:overall} 
\end{figure*}

Existing lifelong knowledge editing methods suffer from catastrophic forgetting, where new edits interfere with prior knowledge, and model collapse, which degrades the model’s original capabilities~\cite{gupta2024rebuildingromeresolving,ma2025perturbationrestrainedsequentialmodelediting,qi2025incontexteditinglearningknowledge}. To address these challenges, retrieval-augmented approaches such as LTE~\cite{jiang2024learningeditaligningllms} and RECIPE~\cite{chen2025lifelongknowledgeeditingllms} are proposed. These methods enhance stability by incorporating retrieved knowledge into models explicitly trained for editing, enabling them to produce revised outputs. 
However, their dataset-specific optimization leads to performance degradation when applied to unseen datasets, necessitating frequent retraining (Figure~\ref{fig:overall}(a)). Furthermore, they impose substantial costs in data requirements, training time, and computational resources (Figure~\ref{fig:overall}(b)). This undermines their scalability and limits their applicability in lifelong learning settings.

To mitigate these issues, we propose \mbox{\textbf{LightEdit}}, a lightweight framework for effective lifelong knowledge editing with minimal training overhead (Figure~\ref{fig:overall}(c)). We hypothesize that, given the edited knowledge in the input context, LLM outputs can be steered by simply suppressing the original probability distribution, without requiring model retraining. To this end, LightEdit employs an edit-aware selector that identifies relevant knowledge requiring modification from retrieved candidates, exerting minimal influence when no edits are needed. Furthermore, we introduce in-context decoding, which suppresses the probability of pre-edit knowledge inferred from contextual cues, thereby guiding the model to integrate the updated knowledge within a single inference pass.

To validate LightEdit, we conduct extensive experiments across lifelong knowledge editing benchmarks using LLaMA-3~\cite{dubey2024llama3herdmodels} and GPT-J~\cite{gpt-j} on the ZSRE~\cite{levy2017zero}, Counterfact~\cite{meng2023locating}, and RIPE~\cite{cohen2023evaluatingrippleeffectsknowledge}. Results demonstrate that LightEdit achieves strong and consistent performance in lifelong settings, offering an efficient solution that preserves the original model parameters while significantly reducing computational overhead. Our approach alleviates dataset dependency and improves scalability, suggesting promising directions for future research in knowledge editing for LLMs. 

Our main contributions are as follows:
\begin{itemize}
    \item We identify the excessive training costs in existing lifelong knowledge editing approaches, highlighting their limited scalability to different datasets.
    \item We propose LightEdit, a lightweight knowledge editing framework that effectively adjusts LLM outputs by minimizing pre-edit probabilities based on selected knowledge.
    \item Through comprehensive experiments on multiple datasets and LLMs, we demonstrate that LightEdit effectively performs lifelong knowledge editing while consuming fewer resources than existing baselines.
\end{itemize}

\section{Related Work}

Knowledge editing aims to correct factual errors encoded in a model with significantly lower computational cost than full retraining~\cite{wang2023knowledge, zhang2024comprehensive}. Existing methods are broadly categorized into parameter-preserving and parameter-modifying approaches. Parameter-preserving methods update model behavior without altering its internal parameters. For instance, SERAC~\cite{mitchell2022memorybased} maintains knowledge edits in an external memory module and adjusts responses accordingly, while IKE~\cite{zheng2023edit} adopts an in-context learning framework to perform edits via prompt manipulation. These approaches mitigate catastrophic forgetting by avoiding parameter updates. Earlier parameter-modifying approaches often employed fine-tuning techniques via multi-loss optimization~\cite{sinitsin2020editable}. However, fine-tuning methods carry risks of overfitting, motivating subsequent developments in hyperparameter-based optimization methods. ROME~\cite{meng2023locating} locates multi-layer perceptrons storing factual knowledge and inserts new key-value pairs into these multi-layer perceptrons to update the model's knowledge. MEMIT~\cite{meng2023massediting} extends ROME by enabling simultaneous edits of substantial amounts of knowledge. Prior research primarily focuses on single or batch editing scenarios, neglecting the incremental toxicity accumulation inherent in sequential parameter modifications. Consequently, these methods have limitations in continuous knowledge editing settings.

Recognizing the need to accommodate multiple sequential edits in realistic scenarios, lifelong editing research has emerged~\cite{hu2024wilkewiselayerknowledgeeditor}. GRACE~\cite{hartvigsen2023aging} maps keys onto the latent space without altering model weights, constructing a localized codebook for effective knowledge editing. R-ROME~\cite{gupta2024rebuildingromeresolving} and PRUNE~\cite{ma2025perturbationrestrainedsequentialmodelediting} address model degradation during sequential edits in existing parameter modification approaches. LTE~\cite{jiang2024learningeditaligningllms} and RECIPE~\cite{chen2025lifelongknowledgeeditingllms} enhance editing capabilities by training models to adapt edits dynamically during inference. RECIPE specifically addresses LTE’s potential overfitting risks by integrating dynamic retrieval mechanisms with p-tuning~\cite{li-liang-2021-prefix,liu-etal-2022-p}. However, these methods either inherit the limitations of prior approaches or demand substantial computational resources, thereby limiting their scalability in practical settings. To overcome these challenges, we propose a lifelong knowledge editing framework that minimizes training overhead while preserving editing effectiveness\footnote{A more in-depth comparison with previous approaches is provided in Appendix~\ref{sec:more_work}.}.

\section{Preliminary}

Knowledge editing aims to update the knowledge encoded in an LLM \( f \) by replacing outdated or incorrect information with accurate content. We represent knowledge as triples \((s, r, o)\), where \(s\) denotes the subject, \(r\) the relation, and \(o\) the object. Given an outdated triple \((s, r, o)\), the editing process replaces the object with a new value \(o^*\), resulting in an updated triple \((s, r, o^*)\). The target set of edited knowledge that the model should reflect is defined as:
\begin{eqnarray}
\mathcal{G}^* = \{(s_i, r_i, o^*_i)\}_{i=1}^{N}.
\end{eqnarray}

In particular, this study does not assume an idealized scenario where all knowledge is simultaneously updated. Instead, we adopt a more realistic lifelong knowledge editing scenario, wherein $N$ knowledge triples in $\mathcal{G}^*$ are sequentially incorporated into the model. Under this setting, an input query $q$ is classified as in-scope if it pertains to any edited knowledge triples $(s_i, r_i, o^*_i)$; otherwise, it is considered out-of-scope. The edited model $f^*$ must satisfy reliability and generality criteria for in-scope queries, and locality for out-of-scope queries. These evaluation criteria are formalized as follows:

\paragraph{Reliability}, a key attribute prioritized in the editing process, assesses whether the model accurately reflects the intended edits. It is defined as:
\begin{eqnarray}
\text{Reliability}(q^{rel}_i) = \mathbb{I}[f^*(q^{rel}_i) = o^*_i],
\end{eqnarray}
where $q^{rel}_i$ is a query explicitly containing the edited knowledge, and $\mathbb{I}$ denotes the indicator function.

\paragraph{Generality} ensures that the edited knowledge remains robustly accessible even when queries are paraphrased. This criterion evaluates the model's ability to output the updated object $o^*_i$ given a paraphrased prompt $q^{gen}_i$. It must satisfy the following condition:
\begin{eqnarray}
\text{Generality}(q^{gen}_i) = \mathbb{I}[f^*(q^{gen}_i) = o^*_i].
\end{eqnarray}

\paragraph{Locality} ensures that edits do not inadvertently disrupt unedited knowledge. The updated model must precisely modify only the targeted information while preserving unrelated knowledge. Hence, the responses to original queries $q^{loc}_j$ must remain unchanged before and after editing:
\begin{eqnarray}
\text{Locality}(q^{loc}_j) = \mathbb{I}[f^*(q^{loc}_j) = f(q^{loc}_j)].
\end{eqnarray}

\section{LightEdit}

The proposed method, LightEdit, is designed to minimize training overhead in lifelong knowledge editing scenarios. When a user provides a new fact \((s, r, o^*)\), it is stored in an external, edited knowledge memory $\mathcal{G}^*$ in natural language form. Upon receiving a natural language query \(q\), relevant entries are retrieved from this memory. To identify which retrieved facts are pertinent for answering the query, we introduce an edit-aware selector, a cross-encoder-based module that assesses the relevance between the query and each candidate knowledge item, filtering out irrelevant entries and retaining only those closely aligned with the query intent. The selected knowledge is then combined with the query and provided as input to the LLM. To incorporate the updated knowledge effectively during inference, we propose in-context decoding, which adjusts the model’s output by explicitly downweighting the log probability of the first token in the original object \(o\). This approach allows the model to suppress outdated knowledge and incorporate new information in a single forward pass, thereby enabling flexible and scalable inference without modifying model parameters.

\subsection{Edit-Aware Selector}

Our approach is inspired by the findings of \citet{zheng2023edit}, which highlight the potential of achieving effective knowledge editing purely through in-context learning. However, their method relies solely on editing-focused prompts, which may compromise the general performance of the underlying LLM. To address this, we propose a selective editing mechanism that allows the unedited model to handle queries unrelated to edited knowledge, while incorporating relevant knowledge only when necessary. Prior work typically selects knowledge based on semantic similarity~\cite{han-etal-2023-improving, chen2025lifelongknowledgeeditingllms}, but such methods often depend on shallow query-knowledge similarity matching and may retrieve irrelevant or omit essential information.

To this end, we employ an edit-aware selector that judges the relevance between a query and retrieved knowledge.
We cast this decision as a natural language inference (NLI) task~\cite{dagan2005pascal,bowman-etal-2015-large}: for each query-knowledge pair, the knowledge item serves as the premise and the query as the hypothesis, and the selector predicts entailment (relevant) vs. non-entailment (irrelevant). 
Concretely, the selector utilizes a pre-trained cross-encoder language model to perform binary classification of the relevance of each query-knowledge pair. The input consists of a user query $q$ and a set of knowledge items $\mathcal{G}^*_{ret}$ returned from an external retrieval system. \( \mathcal{G}^*_{ret} \) comprises the top-\(k\) most relevant facts with respect to \( q \), retrieved from the full edited memory \( \mathcal{G}^* \), and is defined as:
\begin{eqnarray}
\mathcal{G}^*_{ret} = \{(s_i, r_i, o^*_i)\}_{i=1}^{k}.
\end{eqnarray}

The goal is to predict a binary relevance label \( y_i \in \{0, 1\} \) for each candidate knowledge item \((s_i, r_i, o^*_i)\), indicating whether it is pertinent to the query \(q\) and should be included in response generation. Each query–knowledge pair \((q, (s_i, r_i, o^*_i))\) is concatenated into a single input sequence in the format: $\texttt{[CLS]}~q~\texttt{[SEP]}~(s_i, r_i, o^*_i)~\texttt{[SEP]}$, and fed into XLM-RoBERTa~\cite{conneau-etal-2020-unsupervised}. The output embedding corresponding to the \texttt{[CLS]} token, denoted \( h_{\texttt{[CLS]}} \), is used to compute the selection probability \( p_i \) as follows:
\begin{eqnarray}
p_i = \sigma(w^\top h_{\texttt{[CLS]}} + b),
\end{eqnarray}
where \( \sigma(\cdot) \) is the sigmoid activation function, and \( w \) and \( b \) are learnable parameters. The resulting probability \( p_i \) reflects the likelihood that the knowledge item \( (s_i, r_i, o^*_i) \) is relevant to the query. A binary decision \( y_i \) is made by thresholding \( p_i \) at 0.5: if \( y_i = 1 \), the knowledge is included in the selected set \( \mathcal{G}^*_{select} \). If no items are selected (i.e., \( \mathcal{G}^*_{select} = \emptyset \)), the query is processed by the unedited LLM \( f \) without any modification.

The edit-aware selector is trained in a supervised learning, where each query–knowledge pair \((q, (s_i, r_i, o^*_i))\) is annotated with a ground-truth label \(y_i\) indicating whether the knowledge item is relevant to the query. The training minimizes binary cross-entropy loss:
\begin{eqnarray}
\mathcal{L} = - \sum_{i=1}^n [ y_i \log p_i + (1 - y_i) \log(1 - p_i)],
\end{eqnarray}
where $n$ is the number of training samples.

\begin{figure}[t]
\centering 
\includegraphics[width=1\linewidth]{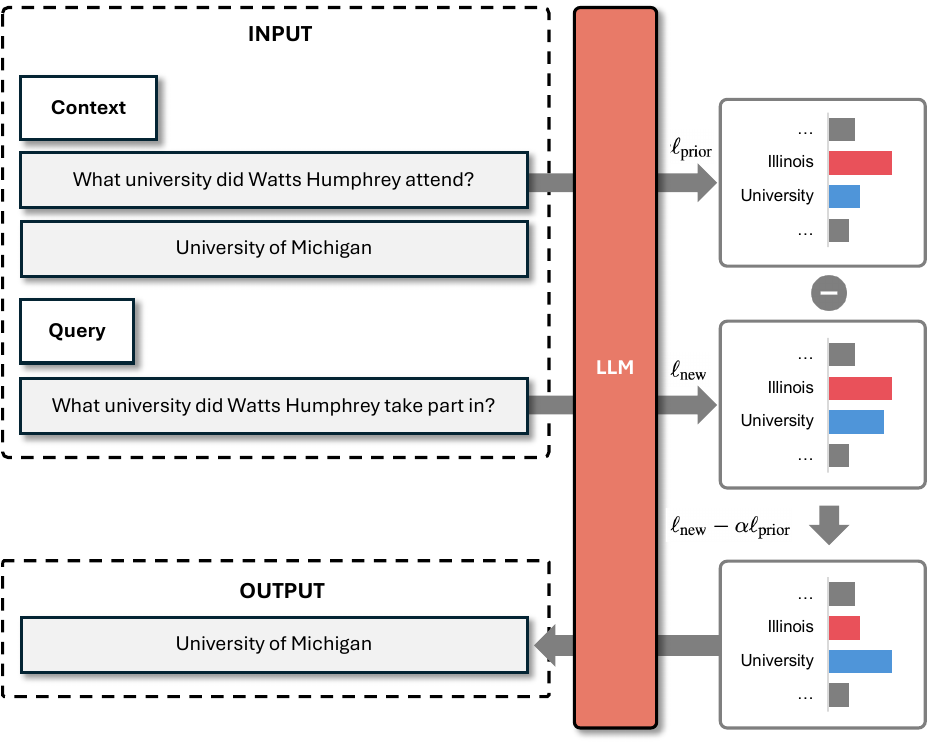}
\caption{An illustration of in-context decoding. This figure depicts a knowledge conflict scenario where the original response (``Illinois Institute of Technology'') has a high probability. In-context decoding suppresses the influence of this prior knowledge, enabling the model to ultimately select the edited information (``University of Michigan'') provided in the context.}
\label{fig:s} 
\end{figure}

\subsection{In-Context Decoding}

As shown in Figure~\ref{fig:s}, this work proposes a non-parametric, probability-based inference approach that dynamically constructs context reflecting new information, enabling more accurate and consistent reasoning in LLMs. Inspired by \citet{li-etal-2023-contrastive}, our method steers model predictions by adjusting contextual information during inference, without retraining model parameters. While the original method aims to enhance target model preferences by capturing likelihood differences across models, our approach instead modulates the generation probability based on the contrast between pre-existing and newly edited knowledge within a single model. This enables dynamic knowledge modification, allowing the model to incorporate new information without conflicting with existing knowledge.

Specifically, we construct a context \(\mathcal{C}\) including the set of new facts $\mathcal{G}^*_{select}$. Given context \(\mathcal{C}\) and query \(q\), the validity of candidate response sequences \(x = x_{1:T}\) is defined by the model's autoregressive probability.
To guide the LLM toward generating responses aligned with the newly edited knowledge, we focus on adjusting the generation probability of the first token. Inspired by \citet{turner2024steeringlanguagemodelsactivation}, we aim to influence the model's response trajectory by modulating the probability distribution of the first token \(x_1\)\footnote{An analysis of the first-token control strategy can be found in \S\ref{sec:first} and Appendix~\ref{sec:embed}.}. Given the context of a factual triple \((s, r)\), the model, based on its pre-trained knowledge, is likely to generate a response corresponding to the original object \(o\), with the first token \(x_o\) reflecting that answer. The generation probability in this case is defined as:
\begin{eqnarray}
\ell_{\text{prior}} = \frac{1}{|\mathcal{G}^*_{select}|} \sum_{(s, r, o^*) \in \mathcal{G}^*_{select}} \log p(x_o \mid s, r).
\end{eqnarray}

In contrast, with the updated context $\mathcal{C}$ and query \(q\), the new context-based generation probability for the first token \(x_1\) is:
\begin{eqnarray}
\ell_{\text{new}} = \log p(x_1 \mid \mathcal{C}, q).
\end{eqnarray}

We encourage the model to favor the newly provided context over its pre-existing knowledge during inference by comparing these two generation probabilities. Specifically, we define the adjusted log probability to guide the model toward newly edited knowledge as:
\begin{eqnarray}
\log p'(x_1) \triangleq \ell_{\text{new}} - \alpha \ell_{\text{prior}},
\end{eqnarray}
where $\alpha$ is a hyperparameter controlling the influence of prior knowledge.

In cases where the updated fact \((s, r, o^*)\) conflicts with the model’s existing knowledge, the generation probability for the original object \(o\) may still remain high. However, by reducing its influence within the new context-based generation probability \(p(x_1 \mid \mathcal{C}, q)\), the model can be flexibly guided to reconstruct its output around the updated knowledge. This approach mitigates knowledge conflicts solely by adjusting the probability distribution of the first token, without requiring any parameter updates. This goes beyond simply producing the correct answer—it demonstrates that an LLM can dynamically restructure its reasoning by revising knowledge.

\section{Experiments}

\begin{table*}[h]
\centering
\resizebox{1\linewidth}{!}{
\begin{tabular}{l|c|ccc|c|ccc|c|ccc}
\toprule
 & \multicolumn{4}{c|}{\textbf{ZSRE}} & \multicolumn{4}{c|}{\textbf{Counterfact}} & \multicolumn{4}{c}{\textbf{RIPE}} \\ \cmidrule(lr){2-5} \cmidrule(lr){6-9} \cmidrule(lr){10-13}
Method & AVG & Reliability & Generality & Locality & AVG & Reliability & Generality & Locality & AVG & Reliability & Generality & Locality \\
\midrule
BASE  & 0.5182 & 0.2845 & 0.2700 & 1.0000 & 0.3377 & 0.0081 & 0.0050 & 1.0000 & 0.5222 & 0.2960 & 0.2706 & 1.0000 \\ \midrule
FT & 0.0916 & 0.1399 & 0.1187 & 0.0161 & 0.0203 & 0.0560 & 0.0050 & 0.0000 & 0.0246 & 0.0352 & 0.0235 & 0.0152 \\
ROME & 0.0363 & 0.0418 & 0.0316 & 0.0354 & 0.0300 & 0.0480 & 0.0420 & 0.0000 & 0.0030 & 0.0058 & 0.0031 & 0.0000 \\
MEMIT & 0.0105 & 0.0000 & 0.0000 & 0.0316 & 0.0013 & 0.0000 & 0.0000 & 0.0040 & 0.0026 & 0.0000 & 0.0000 & 0.0079 \\ 
GRACE & 0.5185 & 0.2851 & 0.2705 & \textbf{1.0000} & 0.3567 & 0.0650 & 0.0050 & \textbf{1.0000} & 0.5474 & 0.3700 & 0.2721 & \textbf{1.0000} \\
R-ROME & 0.0362 & 0.0417 & 0.0315 & 0.0354 & 0.0043 & 0.0030 & 0.0100 & 0.0000 & 0.0084 & 0.0162 & 0.0085 & 0.0005 \\
AlphaEdit & 0.6220 & 0.9143 & 0.5725 & 0.3793 & 0.6295 & 0.7144 & 0.6409 & 0.5333 & 0.5835 & 0.9825 & 0.4820 & 0.2860 \\ \midrule
LTE & 0.7934 & \textbf{0.9772} & 0.7129 & 0.6900 & 0.8753 & 0.9751 & \textbf{0.8556} & 0.7952 & 0.8980 & \textbf{0.9962} & 0.8629 & 0.8350 \\ 
RECIPE & 0.9380 & 0.9226 & 0.9162 & 0.9752 & 0.7664 & 0.8890 & 0.7995 & 0.6108 & 0.5473 & 0.5683 & 0.5053 & 0.5684 \\ \midrule
\cellcolor{gray!20}LightEdit & \cellcolor{gray!20}\textbf{0.9664} & \cellcolor{gray!20}0.9543 & \cellcolor{gray!20}\textbf{0.9483} & \cellcolor{gray!20}0.9966 & \cellcolor{gray!20}\textbf{0.9296} & \cellcolor{gray!20}\textbf{0.9968} & \cellcolor{gray!20}0.8141 & \cellcolor{gray!20}0.9780 & \cellcolor{gray!20}\textbf{0.9818} & \cellcolor{gray!20}0.9848 & \cellcolor{gray!20}\textbf{0.9625} & \cellcolor{gray!20}0.9981 \\
\bottomrule
\end{tabular}
}
\caption{Comparison of methods on ZSRE, Counterfact, and RIPE. The highest score is \textbf{bolded}. AVG denotes the average of the Reliability, Generality, and Locality scores.}
\label{tab:results}
\end{table*}

\subsection{Experimental Setting}

\paragraph{Datasets and Metrics}
We adopt three widely-used evaluation datasets from knowledge editing: ZSRE~\cite{levy2017zero}, Counterfact~\cite{meng2023locating}, and RIPE~\cite{cohen2023evaluatingrippleeffectsknowledge}. Counterfact contains counterfactual knowledge, statements with lower generation probabilities than factual knowledge, provided as new editing knowledge. ZSRE is a context-free question-answering dataset designed for zero-shot relation extraction. RIPE evaluates the ripple effects of injecting new knowledge into related facts. Models are evaluated based on accuracy for reliability, generality, and locality. Further details about the datasets and metrics can be found in Appendix~\ref{sec:appendix_dataset} and Appendix~\ref{sec:appendix_metric}.

\paragraph{Baselines}
The first baseline is the unedited model. FT fine-tunes all parameters of the base model. ROME~\cite{meng2023locating} is an optimization-based method tailored for single-editing tasks, while MEMIT~\cite{meng2023massediting} extends ROME to enable large-scale knowledge editing in a single pass. We incorporate strong baselines specifically designed for lifelong editing scenarios. GRACE~\cite{hartvigsen2023aging} trains only one layer, storing the edited parameters in memory and applying them when an edited query is encountered. R-ROME~\cite{gupta2024rebuildingromeresolving} minimizes model collapse associated with ROME edits and enhances stability during sequential editing. AlphaEdit~\cite{fang2025alphaeditnullspaceconstrainedknowledge} resolves disturbances to existing knowledge caused by the perturbations in locating-then-edit methods using null-space constraints. LTE~\cite{jiang2024learningeditaligningllms}, a retrieval-based lifelong editing method, leverages fine-tuning to learn the knowledge editing task. RECIPE~\cite{chen2025lifelongknowledgeeditingllms} employs p-tuning to mitigate overfitting induced by fine-tuning.

\begin{figure*}[hbt!]
\centering 
\includegraphics[width=0.98\linewidth]{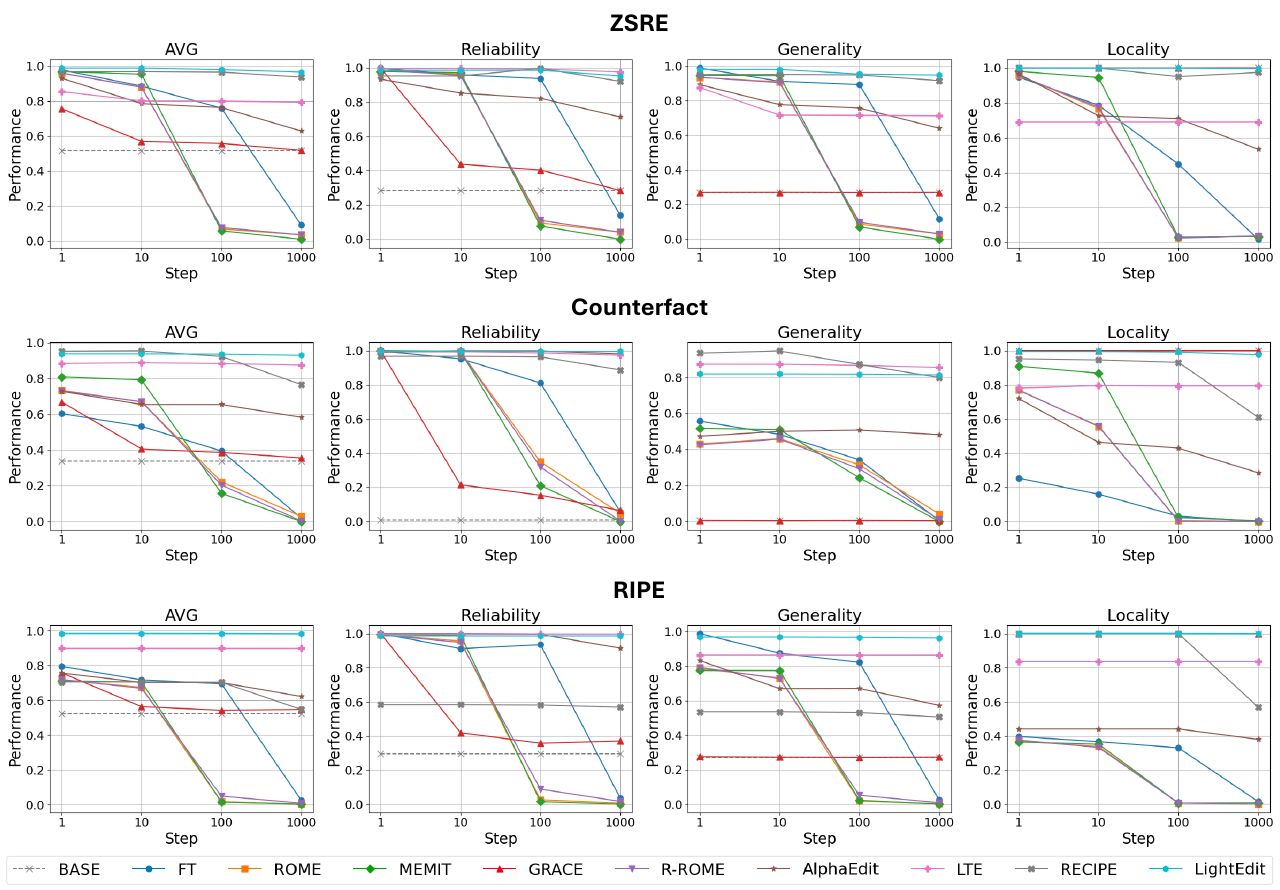}
\caption{Scaling curves that represent editing performance based on the size of edits.}
\label{fig:main} 
\end{figure*}

\paragraph{Implementation Details}
We conduct our experiments using LLaMA-3 (8B)~\cite{dubey2024llama3herdmodels} and GPT-J (6B)~\cite{gpt-j}\footnote{Experiments performed with LLaMA-3 are replicated with GPT-J, detailed in Appendix~\ref{sec:appendix_ex}.}. The model checkpoints used are `meta-llama/Llama-3.1-8B' and `EleutherAI/gpt-j-6B', both accessible via HuggingFace\footnote{\url{https://huggingface.co/}}. 
For the retrieval step, we adopt `multi-qa-mpnet-base-dot-v1'~\cite{reimers-gurevych-2019-sentence} as the retrieval model following \citet{jiang2024learningeditaligningllms}.
For the edit-aware selector, we utilize XLM-RoBERTa-large~\cite{conneau-etal-2020-unsupervised} with the checkpoint `FacebookAI/xlm-roberta-large', trained using the AdamW optimizer with a learning rate of 1e-5, 1 epoch, and batch size of 4. We randomly sample 1,000 instances from the training dataset for training, which is conducted using an NVIDIA A100 GPU. 
The instruction for in-context decoding is provided in Table~\ref{tab:prompt}, and we set $k = 5$ and $\alpha = 0.2$.
For LTE and RECIPE, we follow the training and evaluation protocols, including datasets and settings, as described in their papers. All other baselines are trained and evaluated using the same configurations as EasyEdit~\cite{wang-etal-2024-easyedit}.

\begin{table*}[htbp]
\centering
\resizebox{1\linewidth}{!}{
\begin{tabular}{l|c|ccc|c|ccc|c|ccc}
\toprule

& \multicolumn{4}{c|}{\textbf{ZSRE}} 
& \multicolumn{4}{c|}{\textbf{Counterfact}} 
& \multicolumn{4}{c}{\textbf{RIPE}} \\ \cmidrule(lr){2-5} \cmidrule(lr){6-9} \cmidrule(lr){10-13}
\textbf{$\alpha$} & AVG & Reliability & Generality & Locality 
& AVG & Reliability & Generality & Locality 
& AVG & Reliability & Generality & Locality \\
\midrule
0   & 0.9312 & 0.8978 & 0.8978 & \textbf{0.9979} & 0.8978 & 0.9800 & 0.7305 & 0.9830 & 0.9528 & 0.9520 & 0.9077 & \textbf{0.9987} \\
0.1 & 0.9567 & 0.9366 & 0.9363 & 0.9971 & 0.9198 & 0.9805 & 0.7960 & 0.9830 & 0.9759 & \textbf{0.9857} & 0.9433 & \textbf{0.9987} \\
0.2 & \textbf{0.9664} & \textbf{0.9543} & \textbf{0.9483} & 0.9966 & \textbf{0.9361} & \textbf{0.9972} & \textbf{0.8180} & \textbf{0.9930} & \textbf{0.9818} & 0.9848 & \textbf{0.9625} & 0.9981 \\
0.3 & 0.9644 & 0.9492 & 0.9476 & 0.9964 & 0.8947 & 0.9810 & 0.7250 & 0.9780 & 0.9763 & 0.9747 & 0.9557 & 0.9984 \\
0.4 & 0.9523 & 0.9304 & 0.9300 & 0.9964 & 0.8857 & 0.9830 & 0.6960 & 0.9780 & 0.9609 & 0.9472 & 0.9376 & 0.9978 \\
0.5 & 0.9142 & 0.8687 & 0.8777 & 0.9962 & 0.8732 & 0.9800 & 0.6625 & 0.9770 & 0.9333 & 0.9041 & 0.8979 & 0.9978 \\
\bottomrule
\end{tabular}
}
\caption{Comparison of the effect of the hyperparameter $\alpha$.}
\label{tab:weight}
\end{table*}

\subsection{Main Result}

Table~\ref{tab:results} presents the performance of various methods on 1,000 sequential knowledge editing across the ZSRE, Counterfact, and RIPE.
The FT exhibits a significant decline in reliability and generality, highlighting that straightforward fine-tuning overfits to training data, consequently introducing incorrect edits. Existing editing methods, such as ROME, MEMIT, GRACE, and AlphaEdit, demonstrate specialization in locality or accuracy but fail to achieve high performance across all three metrics. GRACE maintains perfect locality and higher reliability than BASE, yet its limited generality restricts effectiveness for unseen queries. Conversely, ROME and MEMIT, which modify internal model representations locally, fail to generalize these edits effectively, resulting in poor performance in both reliability and generality. LTE and RECIPE, recent retrieval-based editing methods, achieve relatively higher performance. LTE emphasizes reliability and generality at the expense of locality. In contrast, RECIPE demonstrates strong average performance on ZSRE but lacks consistency on RIPE and Counterfact.
Unlike previous methods with clear trade-offs, LightEdit balances the three metrics through selective editing and probabilistic control.
Its two components—the edit-aware selector and in-context decoding—play complementary roles: the former filters relevant knowledge to preserve locality, while the latter adjusts token probabilities to enhance reliability and generality.

Figure~\ref{fig:main} visualizes the performance trends of various methods across different steps. While existing methods exhibit rapid performance degradation in reliability, generality, and locality with increasing steps, LightEdit maintains consistently high initial performance and demonstrates remarkable stability across all metrics, even as the number of editing steps increases. Whereas existing methods excel primarily in specific metrics or benchmarks (e.g., GRACE’s locality or RECIPE’s performance on ZSRE), LightEdit consistently achieves balanced performance across all benchmarks and editing steps. This empirically confirms that the components effectively mitigate the typical trade-offs between reliability, generality, and locality.

\subsection{Analysis}

\paragraph{Optimal Reduction Rate of the Original Knowledge Likelihood.}
Table~\ref{tab:weight} shows the effect of different ratios of prior knowledge probability reduction on editing performance. Generally, increasing the ratio gradually decreased reliability and generality, while locality remained stable across all ratios. Notably, at a ratio of 0.2, the model maintained high average performance and reliability with minimal reduction in generality, achieving average accuracies of 0.9818 and 0.9664 on RIPE and ZSRE, respectively. Based on these results, we set $\alpha=0.2$, confirming that this setting sufficiently improves the model’s ability to reflect edited knowledge without causing excessive knowledge loss.

\begin{figure}[t]
\centering 
\includegraphics[width=0.9\linewidth]{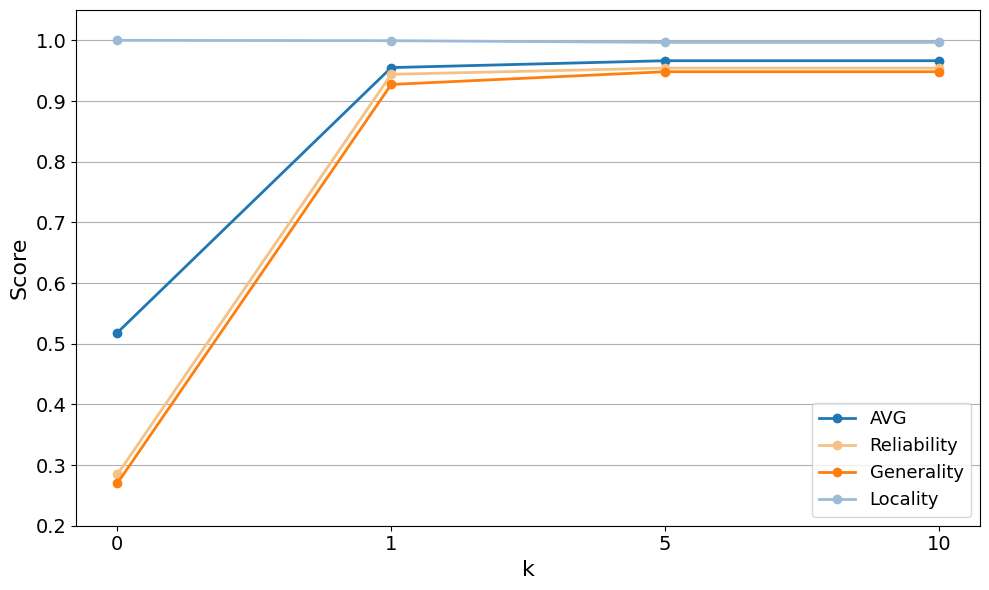}
\caption{Performance variation according to the number of retrieved knowledge.}
\label{fig:topk} 
\end{figure}

\paragraph{Performance Variation by Number of Retrieved Knowledge.}
We conduct additional experiments to analyze the performance variation according to the top-$k$ value in the retrieval process. Experiments are performed with \( k \in \{0, 1, 5, 10\} \), measuring each evaluation metric on ZSRE. Results shown in Figure~\ref{fig:topk} demonstrate significant performance degradation in all metrics when retrieval is not used (\( k=0 \)). Applying retrieval from \( k=1 \) enhances performance across all metrics, quickly reaching near-optimal levels. While a slight performance gain is observed when increasing $k$ to 5, further increasing it to 10 does not yield meaningful additional improvements. Thus, we conclude that setting \( k=5 \) provides optimal performance, and increasing the candidates further does not yield significant additional benefits.

\begin{table}[t]
\centering
\resizebox{0.9\linewidth}{!}{
\begin{tabular}{cc|c|ccc}
\toprule
\textbf{EAS} & \textbf{ICD} & \textbf{AVG} & \textbf{Reliability} & \textbf{Generality} & \textbf{Locality} \\
\midrule
\cellcolor{gray!20}\checkmark & \cellcolor{gray!20}\checkmark & \cellcolor{gray!20}\textbf{0.9664} & \cellcolor{gray!20}\textbf{0.9543} & \cellcolor{gray!20}\textbf{0.9483} & \cellcolor{gray!20}0.9966 \\ \midrule
\checkmark &  & 0.9312 & 0.8978 & 0.8978 & \textbf{0.9979} \\
 & \checkmark & 0.7541 & 0.8898 & 0.9112 & 0.4612 \\
 &  & 0.7615 & 0.9019 & 0.9223 & 0.4603 \\
\bottomrule
\end{tabular}
}
\caption{Ablation of LightEdit components. EAS is Edit-Aware Selector, and ICD is In-Context Decoding.}
\label{tab:ablation}
\end{table}

\begin{table}[t]
\centering
\resizebox{0.9\linewidth}{!}{
\begin{tabular}{l|c|c|c}
\toprule
\textbf{Method} & \textbf{Edit Time} & \textbf{Inference Time} & \textbf{Total Time} \\
\midrule
FT         & \cellcolor{gray!40}0.3594  & \cellcolor{gray!60}0.1847  & \cellcolor{gray!30}0.5441  \\
ROME       & \cellcolor{gray!10}6.0958  & \cellcolor{gray!50}0.1883  & \cellcolor{gray!5}6.2841  \\
MEMIT      & \cellcolor{gray!0}11.7519  & \cellcolor{gray!45}0.1919  & \cellcolor{gray!0}11.9438 \\
GRACE      & \cellcolor{gray!20}4.3634  & \cellcolor{gray!40}0.1939  & \cellcolor{gray!10}4.5573  \\
R-ROME     & \cellcolor{gray!12}5.7078  & \cellcolor{gray!55}0.1897  & \cellcolor{gray!6}5.8975  \\
AlphaEdit  & \cellcolor{gray!0}13.8527  & \cellcolor{gray!65}0.1844  & \cellcolor{gray!0}14.0371 \\
LTE        & \cellcolor{gray!70}0.0000  & \cellcolor{gray!10}0.2111  & \cellcolor{gray!50}0.2111  \\
RECIPE     & \cellcolor{gray!68}0.0099  & \cellcolor{gray!48}0.1944  & \cellcolor{gray!55}0.2043  \\ \midrule
LightEdit  & \cellcolor{gray!70}0.0000  & \cellcolor{gray!20}0.2024  & \cellcolor{gray!60}0.2024  \\
\bottomrule
\end{tabular}
}
\caption{Comparison of edit, inference, and total times measured in seconds. Darker cells indicate better (lower) time.}
\label{tab:edit_times}
\end{table}

\begin{table}[t]
\centering
\resizebox{1\linewidth}{!}{
\begin{tabular}{l|cccc|c}
\toprule
\textbf{Method} & \textbf{CSQA} & \textbf{MMLU} & \textbf{ANLI} & \textbf{TriviaQA} & \textbf{AVG} \\
\midrule
BASE       & 0.7715 & 0.6810 & 0.4661 & 0.5183 & 0.6092 \\ \midrule
FT         & 0.6298 & 0.5518 & 0.4657 & 0.0030 & 0.4126 \\
ROME       & 0.1998 & 0.2686 & 0.3330 & 0.0002 & 0.2004 \\
MEMIT      & 0.1835 & 0.2352 & 0.3521 & 0.0007 & 0.1929 \\
GRACE      & \textbf{0.7715} & \textbf{0.6810} & 0.4661 & \textbf{0.5183} & 0.6092 \\
R-ROME     & 0.2023 & 0.2558 & 0.3270 & 0.0002 & 0.1963 \\
AlphaEdit  & 0.2031 & 0.3656 & 0.3247 & 0.0401 & 0.2334 \\
LTE        & 0.7224 & 0.6171 & 0.3952 & 0.0461 & 0.4452 \\
RECIPE     & \textbf{0.7715} & 0.6805 & 0.4640 & \textbf{0.5183} & 0.6086 \\ \midrule
\cellcolor{gray!20}LightEdit       & \cellcolor{gray!20}\textbf{0.7715} & \cellcolor{gray!20}0.6801 & \cellcolor{gray!20}\textbf{0.4738} & \cellcolor{gray!20}0.5179 & \cellcolor{gray!20}\textbf{0.6108} \\
\bottomrule
\end{tabular}
}
\caption{Performance comparison across multiple LLM datasets.}
\label{tab:performance_comparison}
\end{table}

\paragraph{Ablation Study.}
Table~\ref{tab:ablation} presents the ablation results assessing the contribution of each component.
Removing EAS maintains reliability and generality at relatively high levels but significantly degrades locality, highlighting its key role in filtering query-relevant knowledge.
Removing ICD preserves locality but lowers reliability and generality, indicating its importance for generating reliable and generalized responses.
When both components are removed, performance shows only minor improvement over the ICD-only setting. It remains significantly below that of LightEdit, suggesting that the absence of EAS allows irrelevant information to be incorporated during decoding.
Overall, EAS and ICD are complementary: EAS ensures precise edit localization, while ICD enhances response quality, and their combination yields substantial gains over standard RAG.

\paragraph{Comparison of Editing and Inference Times.}
We analyze editing efficiency by comparing editing, inference, and total times. As shown in Table~\ref{tab:edit_times}, existing editing methods require substantial editing time, with MEMIT and AlphaEdit taking approximately 12 and 14 seconds. In contrast, LightEdit stores knowledge in memory without additional parameter modifications, achieving significantly faster editing. 
Although LightEdit inevitably incurs a slight delay during inference due to the necessity of determining whether editing is required, this step only involves a simple binary classification and thus adds minimal overhead.
These results suggest that LightEdit is efficient for real-world applications that demand on-the-fly knowledge editing.

\paragraph{Performance Comparison Across Various LLM Capability Benchmarks.}
We conduct additional experiments to ensure that knowledge editing does not compromise the various capabilities of LLMs. The evaluation is performed using the LLM harness\footnote{\url{https://github.com/EleutherAI/lm-evaluation-harness}}, and the metric is the F1 score. Further details on the benchmark are given in Appendix~\ref{sec:appendix_dataset}. Table~\ref{tab:performance_comparison} shows significant performance drops after editing for ROME, MEMIT, and AlphaEdit, compared to the base model. In contrast, GRACE and RECIPE maintain performance levels comparable to the base model, while our method slightly improves upon the base model’s performance. These results demonstrate that LightEdit effectively performs knowledge editing without substantially impairing the model’s general capabilities.

\begin{table}[t]
\centering
\resizebox{1\linewidth}{!}{
\begin{tabular}{l|c|ccc}
\toprule
\textbf{Method} & \textbf{AVG} & \textbf{Reliability} & \textbf{Generality} & \textbf{Locality} \\
\midrule
\cellcolor{gray!20}LightEdit (first token) & \cellcolor{gray!20}\textbf{0.9664} & \cellcolor{gray!20}\textbf{0.9543} & \cellcolor{gray!20}\textbf{0.9483} & \cellcolor{gray!20}\textbf{0.9966} \\
All tokens & 0.9174 & 0.8477 & 0.9101 & 0.9944 \\
\bottomrule
\end{tabular}
}
\caption{Comparison between controlling only the first token and all tokens on the ZSRE dataset. }
\label{tab:tokencontrol}
\end{table}

\paragraph{Controlling Only the First Token.} \label{sec:first}

To maintain model fluency while effectively guiding generation, we control the probability distribution of only the first token. 
This strategy leverages the cascade effect, where the first token influences the subsequent generation process. 
Extending control to all tokens was also examined to assess its effect on performance. 
As shown in Table~\ref{tab:tokencontrol}, controlling all tokens resulted in noticeable performance degradation, indicating over-regularization. 
These results suggest that controlling only the first token achieves an effective balance between control strength and fluency.

\section{Conclusion}
In this study, we proposed LightEdit, a knowledge editing framework that enhances the reliability and scalability of LLMs in continually evolving knowledge environments.
Unlike parameter modification-based methods prone to catastrophic forgetting and retrieval-based approaches with high training costs and limited generalizability, LightEdit combines retrieval-based knowledge filtering with probabilistic decoding for efficient lifelong editing.
Our approach minimizes retraining while maintaining high editing accuracy and demonstrates strong scalability across diverse datasets, underscoring its practicality for continuous LLM updates and maintenance.

\section*{Limitations}

While LightEdit demonstrates promising results in lifelong knowledge editing with minimal computational overhead, several limitations remain.

First, the edit-aware selector relies on a pretrained cross-encoder (e.g., XLM-RoBERTa), which requires supervision and additional training. Although this cost is relatively small compared to full fine-tuning approaches, it still presents a barrier in truly zero-shot or resource-constrained environments.
Moreover, as the number of edits increases, the amount of input grows accordingly, which may introduce additional inference overhead during the selection process.

Second, while the proposed approach is evaluated across three benchmark datasets (ZSRE, Counterfact, RIPE), these datasets may not fully represent the diversity of reasoning and factual inconsistency found in complex real-world applications such as open-domain QA, scientific knowledge updates, or multilingual settings.

Lastly, LightEdit’s reliance on adjusting log probabilities during decoding may be sensitive to hyperparameters, such as the suppression coefficient. Although we show robustness at $\alpha=0.2$, tuning this value across domains or models may require nontrivial effort and domain-specific insight.

\section*{Ethical Statement}
This work focuses on improving factual correctness and edit locality in large language models through continual knowledge editing. While our method enhances scalability and reduces training cost, we acknowledge that it does not directly address concerns related to fairness, safety, or the potential misuse of selective knowledge manipulation. The ability to suppress or alter model outputs based on external information raises essential ethical considerations, including the risk of reinforcing biases or enabling automated misinformation. We believe future research should investigate robust safeguards and transparency mechanisms to ensure the responsible deployment of knowledge editing systems in sensitive or high-stakes applications.

\section*{Acknowledgments}

This work was supported by the Commercialization Promotion Agency for R\&D Outcomes (COMPA) grant funded by the Korea government (Ministry of Science and ICT) (2710086166). 
This work was supported by Institute for Information \& communications Technology Promotion (IITP) grant funded by the Korea government (MSIT) (RS-2024-00398115, Research on the reliability and coherence of outcomes produced by Generative AI). 
This work was supported by Institute of Information \& communications Technology Planning \& Evaluation (IITP) under the Leading Generative AI Human Resources Development (IITP-2026-RS-2024-00397085) grant funded by the Korea government (MSIT).
This work was supported by the Ministry of Education of the Republic of Korea and the National Research Foundation of Korea (NRF-2023S1A5C2A07095987).


\bibliography{main}

\begin{thebibliography}{51}
\providecommand{\natexlab}[1]{#1}

\bibitem[{Alabdulmohsin et~al.(2022)Alabdulmohsin, Neyshabur, and Zhai}]{alabdulmohsin2022revisiting}
Ibrahim~M Alabdulmohsin, Behnam Neyshabur, and Xiaohua Zhai. 2022.
\newblock Revisiting neural scaling laws in language and vision.
\newblock \emph{Advances in Neural Information Processing Systems}, 35:22300--22312.

\bibitem[{Bowman et~al.(2015)Bowman, Angeli, Potts, and Manning}]{bowman-etal-2015-large}
Samuel~R. Bowman, Gabor Angeli, Christopher Potts, and Christopher~D. Manning. 2015.
\newblock \href {https://doi.org/10.18653/v1/D15-1075} {A large annotated corpus for learning natural language inference}.
\newblock In \emph{Proceedings of the 2015 Conference on Empirical Methods in Natural Language Processing}, pages 632--642, Lisbon, Portugal. Association for Computational Linguistics.

\bibitem[{Chen et~al.(2025)Chen, Wang, Zhang, Yan, You, Wang, and He}]{chen2025unieditunifiedknowledgeediting}
Qizhou Chen, Dakan Wang, Taolin Zhang, Zaoming Yan, Chengsong You, Chengyu Wang, and Xiaofeng He. 2025.
\newblock \href {https://arxiv.org/abs/2505.12345} {Uniedit: A unified knowledge editing benchmark for large language models}.
\newblock \emph{Preprint}, arXiv:2505.12345.

\bibitem[{Chen et~al.(2024)Chen, Zhang, He, Li, Wang, Huang, and Xue{'}}]{chen2025lifelongknowledgeeditingllms}
Qizhou Chen, Taolin Zhang, Xiaofeng He, Dongyang Li, Chengyu Wang, Longtao Huang, and Hui Xue{'}. 2024.
\newblock \href {https://doi.org/10.18653/v1/2024.emnlp-main.751} {Lifelong knowledge editing for {LLM}s with retrieval-augmented continuous prompt learning}.
\newblock In \emph{Proceedings of the 2024 Conference on Empirical Methods in Natural Language Processing}, pages 13565--13580, Miami, Florida, USA. Association for Computational Linguistics.

\bibitem[{Cohen et~al.(2024)Cohen, Biran, Yoran, Globerson, and Geva}]{cohen2023evaluatingrippleeffectsknowledge}
Roi Cohen, Eden Biran, Ori Yoran, Amir Globerson, and Mor Geva. 2024.
\newblock Evaluating the ripple effects of knowledge editing in language models.
\newblock \emph{Transactions of the Association for Computational Linguistics}, 12:283--298.

\bibitem[{Conneau et~al.(2020)Conneau, Khandelwal, Goyal, Chaudhary, Wenzek, Guzm{\'a}n, Grave, Ott, Zettlemoyer, and Stoyanov}]{conneau-etal-2020-unsupervised}
Alexis Conneau, Kartikay Khandelwal, Naman Goyal, Vishrav Chaudhary, Guillaume Wenzek, Francisco Guzm{\'a}n, Edouard Grave, Myle Ott, Luke Zettlemoyer, and Veselin Stoyanov. 2020.
\newblock \href {https://doi.org/10.18653/v1/2020.acl-main.747} {Unsupervised cross-lingual representation learning at scale}.
\newblock In \emph{Proceedings of the 58th Annual Meeting of the Association for Computational Linguistics}, pages 8440--8451, Online. Association for Computational Linguistics.

\bibitem[{Dagan et~al.(2005)Dagan, Glickman, and Magnini}]{dagan2005pascal}
Ido Dagan, Oren Glickman, and Bernardo Magnini. 2005.
\newblock The pascal recognising textual entailment challenge.
\newblock In \emph{Machine learning challenges workshop}, pages 177--190. Springer.

\bibitem[{Fang et~al.(2025)Fang, Jiang, Wang, Ma, Jie, Wang, He, and seng Chua}]{fang2025alphaeditnullspaceconstrainedknowledge}
Junfeng Fang, Houcheng Jiang, Kun Wang, Yunshan Ma, Shi Jie, Xiang Wang, Xiangnan He, and Tat seng Chua. 2025.
\newblock \href {https://arxiv.org/abs/2410.02355} {Alphaedit: Null-space constrained knowledge editing for language models}.
\newblock \emph{Preprint}, arXiv:2410.02355.

\bibitem[{Gupta et~al.(2024)Gupta, Baskaran, and Anumanchipalli}]{gupta2024rebuildingromeresolving}
Akshat Gupta, Sidharth Baskaran, and Gopala Anumanchipalli. 2024.
\newblock \href {https://doi.org/10.18653/v1/2024.emnlp-main.1210} {Rebuilding {ROME} : Resolving model collapse during sequential model editing}.
\newblock In \emph{Proceedings of the 2024 Conference on Empirical Methods in Natural Language Processing}, pages 21738--21744, Miami, Florida, USA. Association for Computational Linguistics.

\bibitem[{Han et~al.(2023)Han, Li, Tan, Yuanlong, Chai, and Pan}]{han-etal-2023-improving}
Xiaoqi Han, Ru~Li, Hongye Tan, Wang Yuanlong, Qinghua Chai, and Jeff Pan. 2023.
\newblock \href {https://doi.org/10.18653/v1/2023.findings-emnlp.749} {Improving sequential model editing with fact retrieval}.
\newblock In \emph{Findings of the Association for Computational Linguistics: EMNLP 2023}, pages 11209--11224, Singapore. Association for Computational Linguistics.

\bibitem[{Hartvigsen et~al.(2023{\natexlab{a}})Hartvigsen, Sankaranarayanan, Palangi, Kim, and Ghassemi}]{hartvigsen2023aginggracelifelongmodel}
Thomas Hartvigsen, Swami Sankaranarayanan, Hamid Palangi, Yoon Kim, and Marzyeh Ghassemi. 2023{\natexlab{a}}.
\newblock \href {https://arxiv.org/abs/2211.11031} {Aging with grace: Lifelong model editing with discrete key-value adaptors}.
\newblock \emph{Preprint}, arXiv:2211.11031.

\bibitem[{Hartvigsen et~al.(2023{\natexlab{b}})Hartvigsen, Sankaranarayanan, Palangi, Kim, and Ghassemi}]{hartvigsen2023aging}
Tom Hartvigsen, Swami Sankaranarayanan, Hamid Palangi, Yoon Kim, and Marzyeh Ghassemi. 2023{\natexlab{b}}.
\newblock Aging with grace: Lifelong model editing with discrete key-value adaptors.
\newblock \emph{Advances in Neural Information Processing Systems}, 36:47934--47959.

\bibitem[{Haviv et~al.(2023)Haviv, Cohen, Gidron, Schuster, Goldberg, and Geva}]{haviv2023understanding}
Adi Haviv, Ido Cohen, Jacob Gidron, Roei Schuster, Yoav Goldberg, and Mor Geva. 2023.
\newblock \href {https://doi.org/10.18653/v1/2023.eacl-main.19} {Understanding transformer memorization recall through idioms}.
\newblock In \emph{Proceedings of the 17th Conference of the European Chapter of the Association for Computational Linguistics}, pages 248--264, Dubrovnik, Croatia. Association for Computational Linguistics.

\bibitem[{Hendrycks et~al.(2021)Hendrycks, Burns, Basart, Zou, Mazeika, Song, and Steinhardt}]{hendrycks2021measuringmassivemultitasklanguage}
Dan Hendrycks, Collin Burns, Steven Basart, Andy Zou, Mantas Mazeika, Dawn Song, and Jacob Steinhardt. 2021.
\newblock \href {https://arxiv.org/abs/2009.03300} {Measuring massive multitask language understanding}.
\newblock \emph{Preprint}, arXiv:2009.03300.

\bibitem[{Hu et~al.(2024)Hu, Cao, Chen, Liu, and Zhao}]{hu2024wilkewiselayerknowledgeeditor}
Chenhui Hu, Pengfei Cao, Yubo Chen, Kang Liu, and Jun Zhao. 2024.
\newblock \href {https://doi.org/10.18653/v1/2024.findings-acl.207} {{W}il{KE}: Wise-layer knowledge editor for lifelong knowledge editing}.
\newblock In \emph{Findings of the Association for Computational Linguistics: ACL 2024}, pages 3476--3503, Bangkok, Thailand. Association for Computational Linguistics.

\bibitem[{Huang et~al.(2023)Huang, Shen, Zhang, Zhou, Rong, and Xiong}]{huang2023transformerpatchermistakeworthneuron}
Zeyu Huang, Yikang Shen, Xiaofeng Zhang, Jie Zhou, Wenge Rong, and Zhang Xiong. 2023.
\newblock \href {https://arxiv.org/abs/2301.09785} {Transformer-patcher: One mistake worth one neuron}.
\newblock \emph{Preprint}, arXiv:2301.09785.

\bibitem[{Ji et~al.(2023)Ji, Lee, Frieske, Yu, Su, Xu, Ishii, Bang, Madotto, and Fung}]{Ji_2023}
Ziwei Ji, Nayeon Lee, Rita Frieske, Tiezheng Yu, Dan Su, Yan Xu, Etsuko Ishii, Ye~Jin Bang, Andrea Madotto, and Pascale Fung. 2023.
\newblock \href {https://doi.org/10.1145/3571730} {Survey of hallucination in natural language generation}.
\newblock \emph{ACM Computing Surveys}, 55(12):1–38.

\bibitem[{Jiang et~al.(2023)Jiang, Sablayrolles, Mensch, Bamford, Chaplot, Casas, Bressand, Lengyel, Lample, Saulnier et~al.}]{jiang2023mistral}
Albert~Q Jiang, Alexandre Sablayrolles, Arthur Mensch, Chris Bamford, Devendra~Singh Chaplot, Diego de~las Casas, Florian Bressand, Gianna Lengyel, Guillaume Lample, Lucile Saulnier, and 1 others. 2023.
\newblock Mistral 7b.
\newblock \emph{arXiv preprint arXiv:2310.06825}.

\bibitem[{Jiang et~al.(2025)Jiang, Fang, Zhang, Ma, Wan, Wang, He, and seng Chua}]{jiang2025anyediteditknowledgeencoded}
Houcheng Jiang, Junfeng Fang, Ningyu Zhang, Guojun Ma, Mingyang Wan, Xiang Wang, Xiangnan He, and Tat seng Chua. 2025.
\newblock \href {https://arxiv.org/abs/2502.05628} {Anyedit: Edit any knowledge encoded in language models}.
\newblock \emph{Preprint}, arXiv:2502.05628.

\bibitem[{Jiang et~al.(2024)Jiang, Wang, Wu, Zhong, Zeng, Gao, Li, Jiang, Shang, Tang, Liu, and Wang}]{jiang2024learningeditaligningllms}
Yuxin Jiang, Yufei Wang, Chuhan Wu, Wanjun Zhong, Xingshan Zeng, Jiahui Gao, Liangyou Li, Xin Jiang, Lifeng Shang, Ruiming Tang, Qun Liu, and Wei Wang. 2024.
\newblock \href {https://doi.org/10.18653/v1/2024.acl-long.258} {Learning to edit: Aligning {LLM}s with knowledge editing}.
\newblock In \emph{Proceedings of the 62nd Annual Meeting of the Association for Computational Linguistics (Volume 1: Long Papers)}, pages 4689--4705, Bangkok, Thailand. Association for Computational Linguistics.

\bibitem[{Joshi et~al.(2017)Joshi, Choi, Weld, and Zettlemoyer}]{joshi-etal-2017-triviaqa}
Mandar Joshi, Eunsol Choi, Daniel Weld, and Luke Zettlemoyer. 2017.
\newblock \href {https://doi.org/10.18653/v1/P17-1147} {{T}rivia{QA}: A large scale distantly supervised challenge dataset for reading comprehension}.
\newblock In \emph{Proceedings of the 55th Annual Meeting of the Association for Computational Linguistics (Volume 1: Long Papers)}, pages 1601--1611, Vancouver, Canada. Association for Computational Linguistics.

\bibitem[{Lazaridou et~al.(2021)Lazaridou, Kuncoro, Gribovskaya, Agrawal, Liska, Terzi, Gimenez, de~Masson~d'Autume, Kocisky, Ruder et~al.}]{lazaridou2021mind}
Angeliki Lazaridou, Adhi Kuncoro, Elena Gribovskaya, Devang Agrawal, Adam Liska, Tayfun Terzi, Mai Gimenez, Cyprien de~Masson~d'Autume, Tomas Kocisky, Sebastian Ruder, and 1 others. 2021.
\newblock Mind the gap: Assessing temporal generalization in neural language models.
\newblock \emph{Advances in Neural Information Processing Systems}, 34:29348--29363.

\bibitem[{Levy et~al.(2017)Levy, Seo, Choi, and Zettlemoyer}]{levy2017zero}
Omer Levy, Minjoon Seo, Eunsol Choi, and Luke Zettlemoyer. 2017.
\newblock Zero-shot relation extraction via reading comprehension.
\newblock \emph{arXiv preprint arXiv:1706.04115}.

\bibitem[{Li et~al.(2023)Li, Holtzman, Fried, Liang, Eisner, Hashimoto, Zettlemoyer, and Lewis}]{li-etal-2023-contrastive}
Xiang~Lisa Li, Ari Holtzman, Daniel Fried, Percy Liang, Jason Eisner, Tatsunori Hashimoto, Luke Zettlemoyer, and Mike Lewis. 2023.
\newblock \href {https://doi.org/10.18653/v1/2023.acl-long.687} {Contrastive decoding: Open-ended text generation as optimization}.
\newblock In \emph{Proceedings of the 61st Annual Meeting of the Association for Computational Linguistics (Volume 1: Long Papers)}, pages 12286--12312, Toronto, Canada. Association for Computational Linguistics.

\bibitem[{Li and Liang(2021)}]{li-liang-2021-prefix}
Xiang~Lisa Li and Percy Liang. 2021.
\newblock \href {https://doi.org/10.18653/v1/2021.acl-long.353} {Prefix-tuning: Optimizing continuous prompts for generation}.
\newblock In \emph{Proceedings of the 59th Annual Meeting of the Association for Computational Linguistics and the 11th International Joint Conference on Natural Language Processing (Volume 1: Long Papers)}, pages 4582--4597, Online. Association for Computational Linguistics.

\bibitem[{Liu et~al.(2022)Liu, Ji, Fu, Tam, Du, Yang, and Tang}]{liu-etal-2022-p}
Xiao Liu, Kaixuan Ji, Yicheng Fu, Weng Tam, Zhengxiao Du, Zhilin Yang, and Jie Tang. 2022.
\newblock \href {https://doi.org/10.18653/v1/2022.acl-short.8} {{P}-tuning: Prompt tuning can be comparable to fine-tuning across scales and tasks}.
\newblock In \emph{Proceedings of the 60th Annual Meeting of the Association for Computational Linguistics (Volume 2: Short Papers)}, pages 61--68, Dublin, Ireland. Association for Computational Linguistics.

\bibitem[{{Llama Team}(2024)}]{dubey2024llama3herdmodels}
{Llama Team}. 2024.
\newblock \href {https://arxiv.org/abs/2407.21783} {The llama 3 herd of models}.
\newblock \emph{Preprint}, arXiv:2407.21783.

\bibitem[{Ma et~al.(2025)Ma, Wang, Xu, Ling, and Gu}]{ma2025perturbationrestrainedsequentialmodelediting}
Jun-Yu Ma, Hong Wang, Hao-Xiang Xu, Zhen-Hua Ling, and Jia-Chen Gu. 2025.
\newblock \href {https://arxiv.org/abs/2405.16821} {Perturbation-restrained sequential model editing}.
\newblock \emph{Preprint}, arXiv:2405.16821.

\bibitem[{Meng et~al.(2023{\natexlab{a}})Meng, Bau, Andonian, and Belinkov}]{meng2023locating}
Kevin Meng, David Bau, Alex Andonian, and Yonatan Belinkov. 2023{\natexlab{a}}.
\newblock \href {https://arxiv.org/abs/2202.05262} {Locating and editing factual associations in gpt}.
\newblock \emph{Preprint}, arXiv:2202.05262.

\bibitem[{Meng et~al.(2023{\natexlab{b}})Meng, Sharma, Andonian, Belinkov, and Bau}]{meng2023massediting}
Kevin Meng, Arnab~Sen Sharma, Alex Andonian, Yonatan Belinkov, and David Bau. 2023{\natexlab{b}}.
\newblock \href {https://arxiv.org/abs/2210.07229} {Mass-editing memory in a transformer}.
\newblock \emph{Preprint}, arXiv:2210.07229.

\bibitem[{Mitchell et~al.(2022)Mitchell, Lin, Bosselut, Manning, and Finn}]{mitchell2022memorybased}
Eric Mitchell, Charles Lin, Antoine Bosselut, Christopher~D Manning, and Chelsea Finn. 2022.
\newblock Memory-based model editing at scale.
\newblock In \emph{International Conference on Machine Learning}, pages 15817--15831. PMLR.

\bibitem[{Nie et~al.(2020)Nie, Williams, Dinan, Bansal, Weston, and Kiela}]{nie-etal-2020-adversarial}
Yixin Nie, Adina Williams, Emily Dinan, Mohit Bansal, Jason Weston, and Douwe Kiela. 2020.
\newblock \href {https://doi.org/10.18653/v1/2020.acl-main.441} {Adversarial {NLI}: A new benchmark for natural language understanding}.
\newblock In \emph{Proceedings of the 58th Annual Meeting of the Association for Computational Linguistics}, pages 4885--4901, Online. Association for Computational Linguistics.

\bibitem[{OpenAI(2023)}]{openai2023gpt4}
OpenAI. 2023.
\newblock \href {https://arxiv.org/abs/2303.08774} {Gpt-4 technical report}.
\newblock \emph{Preprint}, arXiv:2303.08774.

\bibitem[{Pagnoni et~al.(2021)Pagnoni, Balachandran, and Tsvetkov}]{pagnoni2021understanding}
Artidoro Pagnoni, Vidhisha Balachandran, and Yulia Tsvetkov. 2021.
\newblock \href {https://doi.org/10.18653/v1/2021.naacl-main.383} {Understanding factuality in abstractive summarization with {FRANK}: A benchmark for factuality metrics}.
\newblock In \emph{Proceedings of the 2021 Conference of the North American Chapter of the Association for Computational Linguistics: Human Language Technologies}, pages 4812--4829, Online. Association for Computational Linguistics.

\bibitem[{Pinter and Elhadad(2023)}]{pinter2023emptyingoceanspoonedit}
Yuval Pinter and Michael Elhadad. 2023.
\newblock \href {https://doi.org/10.18653/v1/2023.findings-emnlp.1012} {Emptying the ocean with a spoon: Should we edit models?}
\newblock In \emph{Findings of the Association for Computational Linguistics: EMNLP 2023}, pages 15164--15172, Singapore. Association for Computational Linguistics.

\bibitem[{Qi et~al.(2025)Qi, Yang, Jiang, Wang, Li, Zhong, Yang, and Zheng}]{qi2025incontexteditinglearningknowledge}
Siyuan Qi, Bangcheng Yang, Kailin Jiang, Xiaobo Wang, Jiaqi Li, Yifan Zhong, Yaodong Yang, and Zilong Zheng. 2025.
\newblock \href {https://arxiv.org/abs/2406.11194} {In-context editing: Learning knowledge from self-induced distributions}.
\newblock \emph{Preprint}, arXiv:2406.11194.

\bibitem[{Reimers and Gurevych(2019)}]{reimers-gurevych-2019-sentence}
Nils Reimers and Iryna Gurevych. 2019.
\newblock \href {https://doi.org/10.18653/v1/D19-1410} {Sentence-{BERT}: Sentence embeddings using {S}iamese {BERT}-networks}.
\newblock In \emph{Proceedings of the 2019 Conference on Empirical Methods in Natural Language Processing and the 9th International Joint Conference on Natural Language Processing (EMNLP-IJCNLP)}, pages 3982--3992, Hong Kong, China. Association for Computational Linguistics.

\bibitem[{Sharma et~al.(2024)Sharma, Atkinson, and Bau}]{sharma2024locatingeditingfactualassociations}
Arnab~Sen Sharma, David Atkinson, and David Bau. 2024.
\newblock \href {https://arxiv.org/abs/2404.03646} {Locating and editing factual associations in mamba}.
\newblock \emph{Preprint}, arXiv:2404.03646.

\bibitem[{Shi et~al.(2024)Shi, Xu, Wang, Qin, Wang, Wang, Wang, Ebrahimi, and Wang}]{shi2024continuallearninglargelanguage}
Haizhou Shi, Zihao Xu, Hengyi Wang, Weiyi Qin, Wenyuan Wang, Yibin Wang, Zifeng Wang, Sayna Ebrahimi, and Hao Wang. 2024.
\newblock \href {https://arxiv.org/abs/2404.16789} {Continual learning of large language models: A comprehensive survey}.
\newblock \emph{Preprint}, arXiv:2404.16789.

\bibitem[{Sinitsin et~al.(2020)Sinitsin, Plokhotnyuk, Pyrkin, Popov, and Babenko}]{sinitsin2020editable}
Anton Sinitsin, Vsevolod Plokhotnyuk, Dmitriy Pyrkin, Sergei Popov, and Artem Babenko. 2020.
\newblock Editable neural networks.
\newblock \emph{arXiv preprint arXiv:2004.00345}.

\bibitem[{Talmor et~al.(2019)Talmor, Herzig, Lourie, and Berant}]{talmor-etal-2019-commonsenseqa}
Alon Talmor, Jonathan Herzig, Nicholas Lourie, and Jonathan Berant. 2019.
\newblock \href {https://doi.org/10.18653/v1/N19-1421} {{C}ommonsense{QA}: A question answering challenge targeting commonsense knowledge}.
\newblock In \emph{Proceedings of the 2019 Conference of the North {A}merican Chapter of the Association for Computational Linguistics: Human Language Technologies, Volume 1 (Long and Short Papers)}, pages 4149--4158, Minneapolis, Minnesota. Association for Computational Linguistics.

\bibitem[{Tan et~al.(2024)Tan, Zhang, and Fu}]{tan2024massiveeditinglargelanguage}
Chenmien Tan, Ge~Zhang, and Jie Fu. 2024.
\newblock \href {https://arxiv.org/abs/2311.04661} {Massive editing for large language models via meta learning}.
\newblock \emph{Preprint}, arXiv:2311.04661.

\bibitem[{Turner et~al.(2024)Turner, Thiergart, Leech, Udell, Vazquez, Mini, and MacDiarmid}]{turner2024steeringlanguagemodelsactivation}
Alexander~Matt Turner, Lisa Thiergart, Gavin Leech, David Udell, Juan~J. Vazquez, Ulisse Mini, and Monte MacDiarmid. 2024.
\newblock \href {https://arxiv.org/abs/2308.10248} {Steering language models with activation engineering}.
\newblock \emph{Preprint}, arXiv:2308.10248.

\bibitem[{Wang and Komatsuzaki(2021)}]{gpt-j}
Ben Wang and Aran Komatsuzaki. 2021.
\newblock {GPT-J-6B: A 6 Billion Parameter Autoregressive Language Model}.
\newblock \url{https://github.com/kingoflolz/mesh-transformer-jax}.

\bibitem[{Wang et~al.(2024{\natexlab{a}})Wang, Zhang, Tian, Xi, Yao, Xu, Wang, Mao, Wang, Cheng, Liu, Ni, Zheng, and Chen}]{wang-etal-2024-easyedit}
Peng Wang, Ningyu Zhang, Bozhong Tian, Zekun Xi, Yunzhi Yao, Ziwen Xu, Mengru Wang, Shengyu Mao, Xiaohan Wang, Siyuan Cheng, Kangwei Liu, Yuansheng Ni, Guozhou Zheng, and Huajun Chen. 2024{\natexlab{a}}.
\newblock \href {https://doi.org/10.18653/v1/2024.acl-demos.9} {{E}asy{E}dit: An easy-to-use knowledge editing framework for large language models}.
\newblock In \emph{Proceedings of the 62nd Annual Meeting of the Association for Computational Linguistics (Volume 3: System Demonstrations)}, pages 82--93, Bangkok, Thailand. Association for Computational Linguistics.

\bibitem[{Wang et~al.(2024{\natexlab{b}})Wang, Zhu, Liu, Zheng, Chen, and Li}]{wang2023knowledge}
Song Wang, Yaochen Zhu, Haochen Liu, Zaiyi Zheng, Chen Chen, and Jundong Li. 2024{\natexlab{b}}.
\newblock Knowledge editing for large language models: A survey.
\newblock \emph{ACM Computing Surveys}, 57(3):1--37.

\bibitem[{Xie et~al.(2024)Xie, Cao, Chen, Chen, Liu, and Zhao}]{xie2024memlaenhancingmultilingualknowledge}
Jiakuan Xie, Pengfei Cao, Yuheng Chen, Yubo Chen, Kang Liu, and Jun Zhao. 2024.
\newblock \href {https://arxiv.org/abs/2406.11566} {Memla: Enhancing multilingual knowledge editing with neuron-masked low-rank adaptation}.
\newblock \emph{Preprint}, arXiv:2406.11566.

\bibitem[{Yao et~al.(2023)Yao, Wang, Tian, Cheng, Li, Deng, Chen, and Zhang}]{yao2023editinglargelanguagemodels}
Yunzhi Yao, Peng Wang, Bozhong Tian, Siyuan Cheng, Zhoubo Li, Shumin Deng, Huajun Chen, and Ningyu Zhang. 2023.
\newblock \href {https://doi.org/10.18653/v1/2023.emnlp-main.632} {Editing large language models: Problems, methods, and opportunities}.
\newblock In \emph{Proceedings of the 2023 Conference on Empirical Methods in Natural Language Processing}, pages 10222--10240, Singapore. Association for Computational Linguistics.

\bibitem[{Zhang et~al.(2024)Zhang, Yao, Tian, Wang, Deng, Wang, Xi, Mao, Zhang, Ni, Cheng, Xu, Xu, Gu, Jiang, Xie, Huang, Liang, Zhang, Zhu, Zhou, and Chen}]{zhang2024comprehensive}
Ningyu Zhang, Yunzhi Yao, Bozhong Tian, Peng Wang, Shumin Deng, Mengru Wang, Zekun Xi, Shengyu Mao, Jintian Zhang, Yuansheng Ni, Siyuan Cheng, Ziwen Xu, Xin Xu, Jia-Chen Gu, Yong Jiang, Pengjun Xie, Fei Huang, Lei Liang, Zhiqiang Zhang, and 3 others. 2024.
\newblock \href {https://arxiv.org/abs/2401.01286} {A comprehensive study of knowledge editing for large language models}.
\newblock \emph{Preprint}, arXiv:2401.01286.

\bibitem[{Zhao et~al.(2025)Zhao, Zhou, Li, Tang, Wang, Hou, Min, Zhang, Zhang, Dong, Du, Yang, Chen, Chen, Jiang, Ren, Li, Tang, Liu, Liu, Nie, and Wen}]{zhao2025surveylargelanguagemodels}
Wayne~Xin Zhao, Kun Zhou, Junyi Li, Tianyi Tang, Xiaolei Wang, Yupeng Hou, Yingqian Min, Beichen Zhang, Junjie Zhang, Zican Dong, Yifan Du, Chen Yang, Yushuo Chen, Zhipeng Chen, Jinhao Jiang, Ruiyang Ren, Yifan Li, Xinyu Tang, Zikang Liu, and 3 others. 2025.
\newblock \href {https://arxiv.org/abs/2303.18223} {A survey of large language models}.
\newblock \emph{Preprint}, arXiv:2303.18223.

\bibitem[{Zheng et~al.(2023)Zheng, Li, Dong, Fan, Wu, Xu, and Chang}]{zheng2023edit}
Ce~Zheng, Lei Li, Qingxiu Dong, Yuxuan Fan, Zhiyong Wu, Jingjing Xu, and Baobao Chang. 2023.
\newblock \href {https://doi.org/10.18653/v1/2023.emnlp-main.296} {Can we edit factual knowledge by in-context learning?}
\newblock In \emph{Proceedings of the 2023 Conference on Empirical Methods in Natural Language Processing}, pages 4862--4876, Singapore. Association for Computational Linguistics.

\end{thebibliography}

\appendix

\section{More Related Work}
\label{sec:more_work}

Recent approaches to sequential knowledge editing draw on memory-based methods but typically incur high training costs. We introduce a lightweight framework that reduces these costs while improving reliability, thereby overcoming the limitations of memory-based techniques designed for single edits. For example, IKE~\cite{zheng2023edit} targets the single-edit setting and induces edits via few-shot prompting, which constrains its scalability to iterative or large-scale knowledge updates. In contrast, LightEdit applies training-free in-context decoding at inference time to amplify the probability of edited knowledge while suppressing outdated knowledge, thereby enabling continual editing without requiring model parameter updates. SERAC~\cite{mitchell2022memorybased}, by comparison, requires a separate counterfactual generator and a scope classifier, leading to substantial model overhead and training expense. In contrast, LightEdit operates through training-free probabilistic control of the target LLM, delivering a lightweight framework.

\section{More Implementation Details}
\label{sec:appendix}

\subsection{Datasets Details} \label{sec:appendix_dataset}

We evaluate the overall performance of continual knowledge editing using three benchmark datasets: ZSRE~\cite{levy2017zero}, Counterfact~\cite{meng2023locating}, and RIPE~\cite{cohen2023evaluatingrippleeffectsknowledge}. Table~\ref{tab:dataset_statistics} summarizes key statistics of these datasets. To ensure consistency and fair comparison across settings, we randomly sample 1,000 instances from the test set for evaluation, considering the limited size of the RIPE dataset.

\begin{table}[t]
\centering
\begin{tcolorbox}[colback=gray!10, colframe=black, width=\linewidth, boxrule=0.5pt, arc=2pt]
\{knowledge 1\} 

... 

\{knowledge i\}

\vspace{1em}

Please apply this information to the following sentence instead of the actual facts. You must use this information to answer the following questions with one token.

\vspace{1em}

\{query\}
\end{tcolorbox}
\caption{An example of the instruction used for in-context decoding.}
\label{tab:prompt}
\end{table}

\begin{table}[t]
\centering
\resizebox{0.8\linewidth}{!}{
\begin{tabular}{l|cc}
\toprule
\textbf{Dataset} & \textbf{Train Samples} & \textbf{Test Samples}  \\
\midrule
ZSRE  & 162,555 & 19,009 \\
Counterfact    & 10,000  & 10,000 \\
RIPE  & 3,000   & 1,388   \\
\bottomrule
\end{tabular}
}
\caption{Dataset statistics for ZSRE, Counterfact, and RIPE.}
\label{tab:dataset_statistics}
\end{table}

\begin{table}[t]
\centering
\small
\begin{tabular}{l|ccc}
\toprule
\textbf{Dataset} & \textbf{ZSRE} & \textbf{Counterfact} & \textbf{RIPE} \\
\midrule
\textbf{Accuracy} & 1.0000 & 0.8325 & 0.9616 \\
\bottomrule
\end{tabular}
\caption{Binary classification accuracy of edit-aware selector on each dataset.}
\label{tab:edit_selector_acc}
\end{table}

In addition to assessing the general capabilities of LLMs beyond knowledge editing, we utilize four representative benchmarks:

\paragraph{CSQA} \cite{talmor-etal-2019-commonsenseqa}  
is a multiple-choice benchmark designed to evaluate commonsense reasoning abilities. Questions are based on ConceptNet and typically require general world knowledge and reasoning skills.

\paragraph{MMLU} \cite{hendrycks2021measuringmassivemultitasklanguage}  
includes questions from 57 diverse domains, covering academic and professional topics. Each question is four-choice multiple-choice and evaluates specialized knowledge and problem-solving ability.

\paragraph{ANLI} \cite{nie-etal-2020-adversarial}  
is a natural language inference dataset comprising examples generated by humans specifically to challenge language models. It consists of three rounds (R1, R2, R3) of increasing difficulty and is designed to test a model’s logical reasoning and generalization capabilities.

\paragraph{TriviaQA} \cite{joshi-etal-2017-triviaqa}  
is a large-scale open-domain question answering dataset, built from complex fact-based questions paired with evidence from Wikipedia documents. It measures a model’s ability to perform contextual retrieval and factual answer generation.

\subsection{Metrics Details} \label{sec:appendix_metric}

We follow the evaluation metrics used in ZSRE, Counterfact, and RIPE, as defined by \citet{chen2025lifelongknowledgeeditingllms}. To assess the success of knowledge editing, we adopt three core criteria: reliability, generality, and locality.

\paragraph{Reliability}  
measures whether the edited model produces the correct, updated answer for queries explicitly targeting the modified knowledge. That is, it evaluates whether the intended edit has been successfully reflected in the model’s output. For example, given the query ``Who is the president of Mexico?'' denoted as \( q_i^{rel} \), and assuming the updated answer is ``Claudia Sheinbaum'', the reliability is 1 if the model outputs the correct new object. Formally, reliability is defined as:
\begin{eqnarray}
\frac{1}{|X^{rel}|} \sum_{i=1}^{|X^{rel}|} \mathbb{I}(f^*(q_i^{rel}) = o_i^*),
\end{eqnarray}
where \( X^{rel} \) is the set of queries used for evaluating reliability, \( f^* \) denotes the edited model, and \( o_i^* \) is the correct updated answer.

\begin{table*}[h]
\centering
\resizebox{1\linewidth}{!}{
\begin{tabular}{l|c|ccc|c|ccc|c|ccc}
\toprule
  & \multicolumn{4}{c|}{\textbf{ZSRE}} & \multicolumn{4}{c|}{\textbf{Counterfact}} & \multicolumn{4}{c}{\textbf{RIPE}} \\ \cmidrule(lr){2-5} \cmidrule(lr){6-9} \cmidrule(lr){10-13}
Method & AVG & Reliability & Generality & Locality & AVG & Reliability & Generality & Locality & AVG & Reliability & Generality & Locality \\
\midrule
BASE       & 0.4790 & 0.2215 & 0.2156 & 1.0000 & 0.3351 & 0.0031 & 0.0022 & 1.0000 & 0.4920 & 0.2411 & 0.2348 & 1.0000 \\ \midrule
FT         & 0.3116 & 0.4879 & 0.4129 & 0.0340 & 0.1546 & 0.3275 & 0.1175 & 0.0189 & 0.4792 & 0.6443 & 0.5855 & 0.2077 \\
ROME       & 0.0120 & 0.0191 & 0.0170 & 0.0000 & 0.0003 & 0.0010 & 0.0000 & 0.0000 & 0.0046 & 0.0054 & 0.0031 & 0.0054 \\
MEMIT      & 0.8005 & 0.8534 & 0.7927 & 0.7555 & 0.6787 & 0.9579 & 0.4460 & 0.6323 & 0.4251 & 0.5229 & 0.3418 & 0.4107 \\
GRACE      & 0.5249 & 0.3586 & 0.2161 & \textbf{1.0000} & 0.3558 & 0.0640 & 0.0033 & \textbf{1.0000} & 0.5160 & 0.3108 & 0.2372 & \textbf{1.0000} \\
R-ROME     & 0.0404 & 0.0156 & 0.0148 & 0.0909 & 0.0510 & 0.0640 & 0.0890 & 0.0000 & 0.0049 & 0.0063 & 0.0058 & 0.0025 \\ 
AlphaEdit & 0.6597 & \textbf{0.9943} & 0.6621 & 0.3225 & 0.8563 & 0.9307 & 0.7477 & 0.8906 & 0.7307 & 0.9930 & 0.4090 & 0.7900 \\ \midrule
LTE        & 0.7831 & 0.9819 & 0.6978 & 0.6695 & 0.8787 & 0.9775 & \textbf{0.8770} & 0.7815 & 0.9007 & \textbf{0.9976} & 0.8748 & 0.8296 \\
RECIPE     & 0.9612 & 0.9459 & 0.9377 & \textbf{1.0000} & 0.7482 & 0.7565 & 0.6830 & 0.8050 & 0.4877 & 0.5813 & 0.4834 & 0.3983 \\ \midrule
\cellcolor{gray!20}LightEdit      & \cellcolor{gray!20}\textbf{0.9776} & \cellcolor{gray!20}0.9939 & \cellcolor{gray!20}\textbf{0.9405} & \cellcolor{gray!20}0.9984 & \cellcolor{gray!20}\textbf{0.8993} & \cellcolor{gray!20}\textbf{0.9910} & \cellcolor{gray!20}0.7320 & \cellcolor{gray!20}0.9750  & \cellcolor{gray!20}\textbf{0.9864} & \cellcolor{gray!20}0.9942 & \cellcolor{gray!20}\textbf{0.9661} & \cellcolor{gray!20}0.9988 \\
\bottomrule
\end{tabular}
}
\caption{Comparison of methods on ZSRE, Counterfact, and RIPE using GPT-J.}
\label{tab:main_gpt}
\end{table*}

\paragraph{Generality}  
assesses whether the edit is consistently reflected across semantically equivalent variations of the edited query. In other words, the model should continue to generate the correct updated answer even when the question is rephrased. For example, given a paraphrased query \( q_i^{gen} \), such as ``Who currently leads Mexico?'', the model should still return the updated answer ``Claudia Sheinbaum'' to achieve high Generality. This metric is formally defined as:
\begin{eqnarray}
\frac{1}{|X^{gen}|} \sum_{i=1}^{|X^{gen}|} \mathbb{I}(f^*(q_i^{gen}) = o_i^*),
\end{eqnarray}
where \( X^{gen} \) denotes the set of paraphrased queries used to evaluate generality.

\paragraph{Locality}  
evaluates whether the edit has left unrelated knowledge intact. For queries unrelated to the edited fact, the model’s responses should remain unchanged before and after the edit. For example, if the query is ``What is the capital of France?''—a fact that has not been edited—the model should continue to answer ``Paris'' both before and after the edit. Any deviation indicates a violation of Locality. The metric is formally defined as:
\begin{eqnarray}
\frac{1}{|X^{loc}|} \sum_{i=1}^{|X^{loc}|} \mathbb{I}(f^*(q_i^{loc}) = f(q_i^{loc})),
\end{eqnarray}
where \( X^{loc} \) denotes the set of locality-checking queries, \( f \) is the original model, and \( f^* \) is the edited model.

\begin{figure*}[hbt!]
\centering 
\includegraphics[width=1\linewidth]{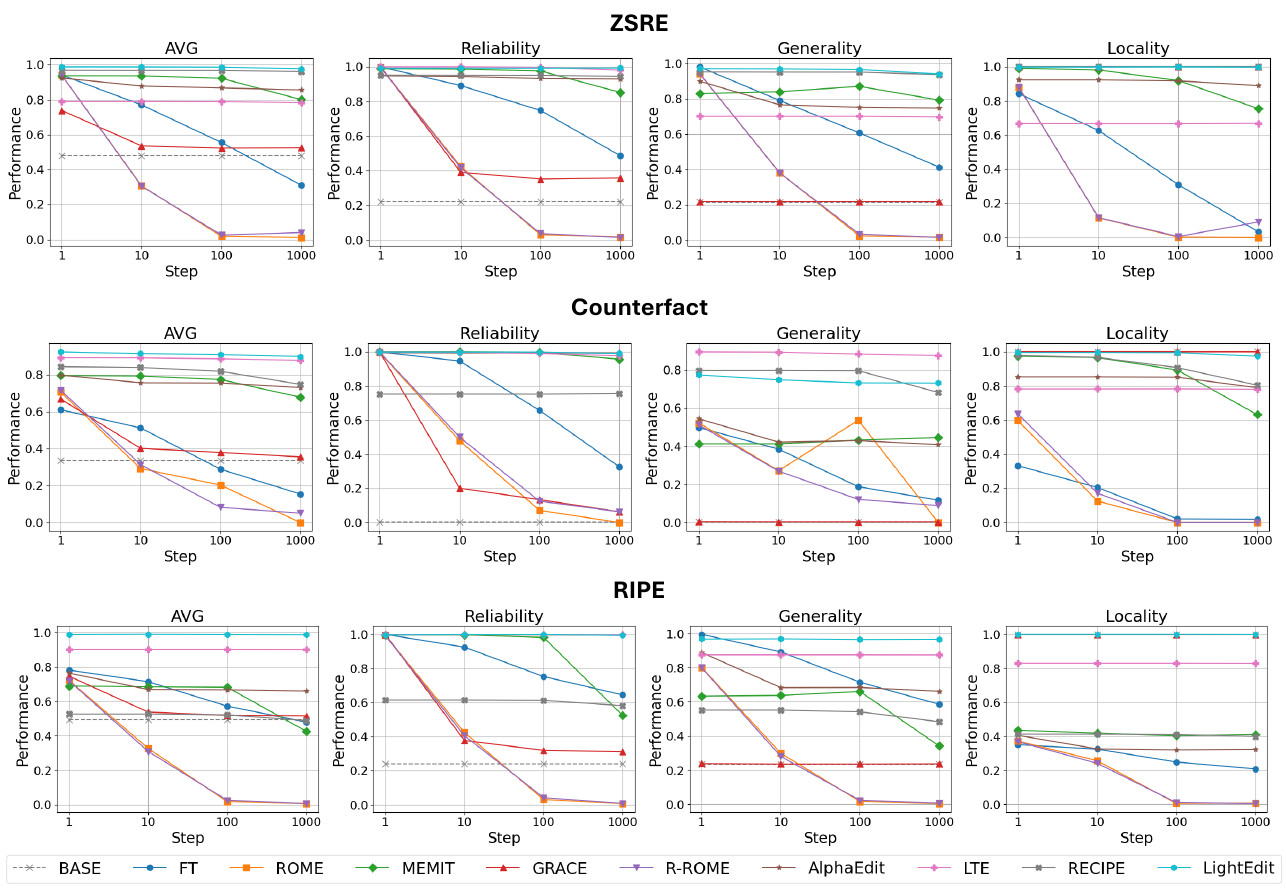}
\caption{Scaling curves that represent editing performance based on the size of edits using GPT-J.}
\label{fig:main_gpt} 
\end{figure*}

\begin{table*}[htbp]
\centering
\resizebox{1\linewidth}{!}{
\begin{tabular}{c|c|ccc|c|ccc|c|ccc}
\toprule

& \multicolumn{4}{c|}{\textbf{ZSRE}} 
& \multicolumn{4}{c|}{\textbf{Counterfact}} 
& \multicolumn{4}{c}{\textbf{RIPE}} \\ \cmidrule(lr){2-5} \cmidrule(lr){6-9} \cmidrule(lr){10-13}
$\alpha$ & AVG & Reliability & Generality & Locality 
& AVG & Reliability & Generality & Locality 
& AVG & Reliability & Generality & Locality \\
\midrule
0.0  & 0.8473 & 0.9420 & 0.6190 & 0.9810 & 0.9145 & 0.9442 & 0.8013 & 0.9981 & 0.9556 & 0.9313 & 0.9368 & \textbf{0.9988} \\
0.1  & 0.8897 & 0.9820 & 0.7080 & \textbf{0.9790} & 0.9657 & 0.9883 & 0.9107 & 0.9981 & \textbf{0.9880} & \textbf{0.9981} & \textbf{0.9672} & \textbf{0.9988} \\
0.2  & 0.8993 & \textbf{0.9910} & 0.7320 & 0.9750  & \textbf{0.9776} & \textbf{0.9939} & \textbf{0.9405} & \textbf{0.9984} & 0.9864 & 0.9942 & 0.9661 & \textbf{0.9988} \\
0.3  & \textbf{0.9027} & 0.9900 & \textbf{0.7430} & 0.9750 & 0.9708 & 0.9877 & 0.9273 & 0.9975 & 0.9790 & 0.9827 & 0.9557 & 0.9985 \\
0.4  & 0.8967 & \textbf{0.9910} & 0.7240 & 0.9750 & 0.9505 & 0.9688 & 0.8865 & 0.9962 & 0.9540 & 0.9314 & 0.9329 & 0.9977 \\
0.5  & 0.8883 & 0.9860 & 0.7040 & 0.9750 & 0.9155 & 0.9224 & 0.8281 & 0.9959 & 0.9031 & 0.8342 & 0.8775 & 0.9977 \\
\bottomrule
\end{tabular} }
\caption{Comparison of the effect of the hyperparameter $\alpha$ using GPT-J.}
\label{tab:weight_gpt}
\end{table*}

\subsection{Fine-tuning Details of Edit-Aware Selector}

Edit-aware selector is a binary classifier based on XLM-RoBERTa, fine-tuned to determine whether a given knowledge edit candidate should be applied. For fine-tuning, we randomly sampled a total of 1,000 instances from the publicly available training sets of ZSRE, Counterfact, and RIPE. All samples were selected in a reproducible manner with a fixed random seed (random\_state = 42), and no overlap was allowed between training and evaluation samples. As shown in Table~\ref{tab:edit_selector_acc}, the model performance was evaluated on 1,000 unseen samples from each dataset, using binary classification accuracy as the evaluation metric. The perfect accuracy on ZSRE suggests that the signal for edit applicability is relatively clear in this dataset. In contrast, Counterfact shows lower accuracy due to the greater diversity of factual inversion cases, while RIPE maintains high performance. These results confirm that our module can consistently distinguish whether an edit should be applied across diverse knowledge editing datasets.

\section{More Experimental Results} \label{sec:appendix_ex}

\subsection{Main Result using GPT-J}

In this section, we comprehensively compare and analyze the performance of various knowledge editing methods using GPT-J as the base model. Table~\ref{tab:main_gpt} reports the average scores and the three core evaluation metrics, while Figure~\ref{fig:main_gpt} visualizes how each method’s performance evolves across editing steps, highlighting their robustness and generalization capabilities.

As shown in Table~\ref{tab:main_gpt}, LightEdit achieves the highest average performance across all benchmarks and metrics. Notably, it consistently maintains a locality score above 0.99 on all datasets, indicating minimal interference with unedited knowledge.

This trend is further supported by Figure~\ref{fig:main_gpt}, where LightEdit demonstrates remarkable stability across all editing steps. In contrast, existing methods such as ROME, FT, and MEMIT show steep declines in reliability and generality as the number of editing steps increases. For instance, in ZSRE, LightEdit sustains its initial reliability throughout the sequence, whereas other methods either start with low performance or deteriorate rapidly over time.

Moreover, while certain baselines specialize in isolated metrics, such as GRACE achieving perfect locality or RECIPE being highly optimized for ZSRE, LightEdit uniquely balances all three metrics, effectively overcoming their typical trade-offs. This is attributed to its core components: the edit-aware selector, which ensures precise knowledge targeting, and in-context decoding, which preserves original knowledge without parameter updates. These results clearly demonstrate that LightEdit offers superior accuracy and robustness compared to parameter-modifying and retrieval-based approaches.

\subsection{Ablation Study on α using GPT-J}

Table~\ref{tab:weight_gpt} presents a quantitative analysis of how the editing performance of LightEdit varies with different values of the key hyperparameter $\alpha$, which controls the degree of suppression applied to the original knowledge probability. As $\alpha$ increases, the suppression generally becomes stronger. Overall, $\alpha=0.2$ yields the best balance across all three benchmarks, demonstrating LightEdit's ability to effectively manage trade-offs between metrics through probability-based control. In contrast, setting $\alpha$ too low or too high tends to degrade performance in either generality or reliability, highlighting the critical role of proper hyperparameter tuning in achieving optimal performance.

\begin{figure*}[hbt!]
\centering 
\includegraphics[width=1\linewidth]{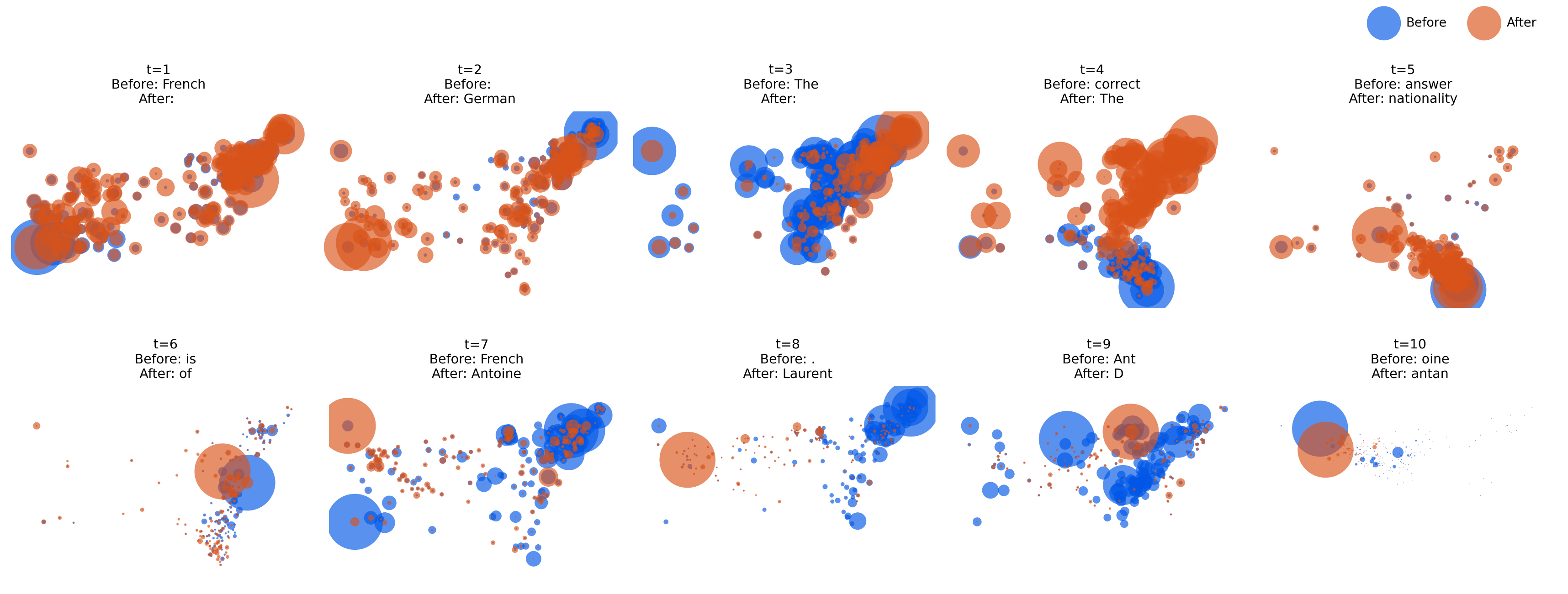}
\caption{Visualization of next-token distributions before and after applying in-context decoding. Each panel corresponds to one generated token ($t=1,\ldots,10$), where blue points represent the \textcolor{blue}{Before} distribution and orange points represent the \textcolor{orange}{After} distribution. Point size is proportional to the square root of token probability, and the title of each panel indicates the Top-1 token for both sides (Before / After).}
\label{fig:embed} 
\end{figure*}

\subsection{Semantic Space Shift Before and After In-Context Decoding}
\label{sec:embed}

Figure~\ref{fig:embed} visualizes how the token probability distributions change before and after applying in-context decoding. 
For each generated token $t$, we extract the embeddings of the top-10 most probable tokens from the model’s embedding space, 
project them into two dimensions using PCA. In this experiment, we edited the fact in the prompt \textit{``The nationality of Antoine Laurent Dantan is?''} to set the new target as \textbf{German}. 
We then queried a semantically different expression, \textit{``Is Antoine Laurent Dantan’s nationality?''}, to evaluate the model’s generalization after editing. Without in-context decoding, the model failed to incorporate the updated knowledge and continued to output the original answer, \textbf{French}.
The embedding distributions show that, at early decoding steps, the probability mass is widely dispersed and does not focus toward the correct candidate. 
In contrast, with in-context decoding applied, the center of the distribution shifts toward ``German'' from the very first token step. 
This indicates that in-context decoding realigns the semantic space at the first decoding step, 
allowing the updated knowledge to be directly reflected in the subsequent token prediction process. 
In summary, in-context decoding does not merely adjust token probabilities; 
It modifies the embedding distribution itself at the first token step, 
guiding the model to integrate the newly edited knowledge semantically and converge to the correct answer.

\begin{table}[t]
\centering
\resizebox{0.85\linewidth}{!}{
\begin{tabular}{l|c|ccc}
\toprule
\textbf{\#Edits} & \textbf{AVG} & \textbf{Reliability} & \textbf{Generality} & \textbf{Locality} \\
\midrule
5{,}000 & 0.9633 & 0.9538 & 0.9417 & 0.9944 \\
1{,}000 & 0.9664 & 0.9543 & 0.9483 & 0.9966 \\
\bottomrule
\end{tabular}
}
\caption{Performance comparison when increasing the number of edits (\#Edits). }
\label{tab:newfacts}
\end{table}

\subsection{Performance Degradation When Increasing the Number of Edits}

Our proposed method shows minimal performance degradation even when a large number of edits are applied, substantially reducing the need for retraining. 
Table~\ref{tab:newfacts} reports the results as the number of edits increases up to 5,000. 
As shown, the performance drop is negligible, demonstrating that the model maintains stable accuracy and low sensitivity to the scale of applied edits.

\begin{table}[t]
\centering
\resizebox{1\linewidth}{!}{
\begin{tabular}{l|c|c|ccc}
\toprule
\textbf{Method} & \textbf{\#Edits} & \textbf{AVG} & \textbf{Reliability} & \textbf{Generality} & \textbf{Locality} \\
\midrule
\cellcolor{gray!20}LightEdit & \cellcolor{gray!20}1  & \cellcolor{gray!20}\textbf{0.9892} & \cellcolor{gray!20}\textbf{0.9872} & \cellcolor{gray!20}\textbf{0.9809} & \cellcolor{gray!20}\textbf{0.9994} \\
\cellcolor{gray!20}LightEdit & \cellcolor{gray!20}10 & \cellcolor{gray!20}\textbf{0.9892} & \cellcolor{gray!20}\textbf{0.9872} & \cellcolor{gray!20}\textbf{0.9809} & \cellcolor{gray!20}\textbf{0.9994} \\
\midrule
IKE & 1  & 0.8342 & 0.9980 & 0.8487 & 0.6559 \\
IKE & 10 & 0.8236 & 0.9969 & 0.7789 & 0.6949 \\
\bottomrule
\end{tabular}
}
\caption{Comparison between IKE and LightEdit on the ZSRE dataset under limited context length.}
\label{tab:ike}
\end{table}

\subsection{Comparison with IKE} \label{sec:ike}

IKE~\cite{zheng2023edit} is an in-context learning-based knowledge editing method that performs single-edit operations by injecting new information into the prompt. 
Our approach differs in both scalability and mechanism. 
IKE relies on long input contexts to represent new knowledge, which inherently limits its applicability when the number of edits increases. 
In contrast, LightEdit performs lightweight editing through explicit edit-aware selection and controlled decoding, enabling efficient handling of up to 1,000 edits without prompt length constraints. To further examine this distinction, we evaluated IKE under restricted context settings on the ZSRE dataset. 
As shown in Table~\ref{tab:ike}, IKE demonstrates lower generality and locality compared to LightEdit, indicating that few-shot in-context editing struggles to maintain accuracy and consistency when scaling to multiple edits. 
These results highlight that LightEdit achieves scalable and stable knowledge editing beyond the limitations of ICL-based methods.

\subsection{Evaluation on UniEdit Benchmark}
\label{appendix:uniedit}

We further evaluate LightEdit on the UniEdit benchmark~\cite{chen2025unieditunifiedknowledgeediting}, which provides a broader testbed covering diverse domains and evaluation criteria. Table~\ref{tab:uniedit_results} reports the results.
LightEdit achieves the best overall average performance among all compared methods, obtaining an average score of 91.65.
In particular, LightEdit shows consistently strong and well-balanced performance across all three editing metrics.
We emphasize that these results are out-of-distribution evaluations.
The edit-aware selector used in LightEdit was trained only on ZSRE, Counterfact, and RIPE, without any training on UniEdit.
Despite this setting, LightEdit still generalizes well and outperforms all baselines in terms of average performance.

\begin{table}[t]
\centering
\resizebox{1\linewidth}{!}{
\begin{tabular}{l|c|ccc}
\toprule
\textbf{Method} & \textbf{AVG} & \textbf{Reliability} & \textbf{Generality} & \textbf{Locality} \\
\midrule
BASE        & 65.16 & 43.68 & 51.81 & 100.00 \\ \midrule
FT          & 87.51 & \textbf{100.00} & 69.00 & 93.54 \\
IKE         & 87.95 & 93.54 & \textbf{89.52} & 80.79 \\
ROME        & 74.10 & 75.81 & 51.38 & 95.12 \\
SERAC       & 88.96 & 98.96 & 85.66 & 84.25 \\
T-Patcher   & 68.71 & 73.03 & 49.83 & 83.27 \\
GRACE       & 83.93 & 99.92 & 51.89 & \textbf{99.97} \\
AlphaEdit   & 79.30 & 84.09 & 55.10 & 98.72 \\ \midrule
\cellcolor{gray!20}LightEdit  & \cellcolor{gray!20}\textbf{91.65} & \cellcolor{gray!20}99.32 & \cellcolor{gray!20}83.27 & \cellcolor{gray!20}92.36 \\
\bottomrule
\end{tabular}
}
\caption{Evaluation Results on UniEdit.}
\label{tab:uniedit_results}
\end{table}

\begin{table}[t]
\centering
\renewcommand{\arraystretch}{1}
\setlength{\tabcolsep}{6pt}
\footnotesize

\begin{tabular}{p{1.48cm}|p{5.3cm}
}
\toprule
\multicolumn{2}{c}{\cellcolor{gray!20}\textbf{Generality Example}} \\ 
\midrule
Question & What was Salvatore Papaccio's range like? \\ \midrule
Edited Knowledge & What type of tone does Salvatore Papaccio sing in? → bass \\ \midrule
Target \newline Answer & 
bass  \\
\midrule
RECIPE & \cellcolor{red!15}\texttt{soprano} \\
LightEdit & \cellcolor{green!20}\texttt{bass} \\

\midrule[0.8pt]
\multicolumn{2}{c}{\cellcolor{gray!20}\textbf{Locality Example}} \\ 
\midrule
Question & ASP.NET MVC Framework was created by \\ \midrule
Edited Knowledge & 
--  \\ \midrule
Target \newline Answer & 
Microsoft  \\ 
\midrule
RECIPE & \cellcolor{red!15}\texttt{Netherlands} \\
LightEdit & \cellcolor{green!20}\texttt{Microsoft} \\

\bottomrule
\end{tabular}

\caption{
Case study illustrating the difference between RECIPE and LightEdit. 
}
\label{tab:case}
\end{table}

\subsection{Case Study}

Table~\ref{tab:case} presents two representative examples that highlight the behavioral differences between RECIPE and LightEdit, in terms of generality and locality. 
In the Generality example, the retrieved evidence correctly contains the updated knowledge. 
However, even though RECIPE successfully retrieves this information, it fails to reflect the newly edited knowledge in its final response, continuing to output the outdated answer \texttt{soprano}. 
In contrast, LightEdit incorporates the revised knowledge and correctly answers \texttt{bass}, demonstrating its ability to integrate edited information while maintaining general consistency.
In the Locality example, the retrieved corpus includes irrelevant information -- 
\textit{``The headquarter of WaterAid is located in → Netherlands''}. 
RECIPE fails to filter out this unrelated content and incorrectly generates the answer \texttt{Netherlands}. 
By comparison, LightEdit effectively suppresses such noisy retrievals and outputs the correct answer \texttt{Microsoft}, showing improved locality and robustness to spurious retrieved facts.

\end{document}